\documentclass{article}

\usepackage[main, final]{neurips_2026}

\usepackage[utf8]{inputenc}
\usepackage[T1]{fontenc}
\usepackage{hyperref}
\usepackage{url}
\usepackage{booktabs}
\usepackage{multirow}
\usepackage{amsfonts}
\usepackage{nicefrac}
\usepackage{microtype}
\usepackage[table]{xcolor}

\usepackage{graphicx}
\usepackage{subcaption}
\usepackage{wrapfig}
\usepackage{placeins}
\usepackage{float}

\graphicspath{{../images/}{images/}}

\usepackage{amsmath}
\usepackage{amssymb}
\usepackage{mathtools}
\usepackage{amsthm}

\usepackage{algorithm}
\usepackage{algorithmic}

\newcommand{\vx}{\boldsymbol{x}}
\newcommand{\vh}{\boldsymbol{h}}
\newcommand{\mask}{\texttt{M}}

\makeatletter
\newcommand{\STATEx}{\item[]}

\makeatother

\usepackage[capitalize,noabbrev]{cleveref}

\theoremstyle{plain}

\theoremstyle{definition}

\theoremstyle{remark}

\usepackage[textsize=tiny]{todonotes}

\usepackage{tikz}
\usetikzlibrary{decorations.pathreplacing}

\usepackage[most]{tcolorbox}
\definecolor{takeawayblue}{HTML}{1F77B4}
\newtcolorbox{takeaway}{enhanced, sharp corners, boxrule=0pt, frame hidden,
  colback=takeawayblue!10, borderline west={2pt}{0pt}{takeawayblue},
  left=6pt, right=6pt, top=4pt, bottom=4pt}

\title{Re-evaluating Confidence Remasking in \\ Masked Diffusion Language Models}

\author{%
  Stipe Frković$^{1,}$\thanks{Equal contribution. Corresponding authors: $<$ \texttt{stipe.frkovic@student.uva.nl}, \texttt{m.jazbec@uva.nl}$>$} \quad Metod Jazbec$^{1,*}$ \quad \textbf{Dan Zhang}$^2$ \\
  \textbf{Christian A. Naesseth}$^1$ \quad \textbf{Ilija Bogunovic}$^{3}$ \quad \textbf{Eric Nalisnick}$^{4}$\\
  $^1$UvA-Bosch Delta Lab, University of Amsterdam  \quad 
  $^2$Bosch Center for AI \\
  $^3$University of Basel  \quad 
  $^4$Johns Hopkins University
}

\begin{document}

\maketitle

\begin{abstract}
  Masked diffusion language models (dLLMs) have recently emerged as a competitive alternative to autoregressive language models, with the promise of faster inference via parallel token generation. A notable limitation of the masked formulation, however, is that once a token has been unmasked it can no longer be revised, leaving dLLMs vulnerable to early sampling mistakes. Confidence-based remasking methods aim to address this in a training-free manner by revising low-confidence tokens, with encouraging early results. In this work, we compare the accuracy--compute Pareto frontiers of three such methods (WINO, Saber, and COVER) to those of confidence-based unmasking alone (Fast-dLLM). We find that under greedy decoding, all three perform on par with Fast-dLLM at commonly used shorter block lengths; gains of remasking appear only at longer block lengths, where confidence-based unmasking is known to degrade. Under stochastic decoding, WINO reduces generation diversity beyond that of unmasking alone, while it helps recover from errors introduced by learned Bernoulli unmasking policies. Our results show that the benefits of confidence-based remasking are highly setting-dependent, underscoring the need for more comprehensive evaluation. Finally, our analyses point to the dLLM, not the remasking rule, as the bottleneck, suggesting that effective remasking may require changes to training, not only to inference.
\end{abstract}

\section{Introduction}
Autoregressive (AR) language models have become the dominant paradigm for text generation, but their sequential left-to-right decoding inherently limits inference throughput. Masked diffusion language models (dLLMs; \citealt{sahoo2024simple,shi2024simplified}) have recently emerged as a competitive alternative: starting from a fully masked sequence, dLLMs progressively unmask tokens in arbitrary order, enabling parallel decoding within each step. Recent work has scaled dLLMs to billions of parameters \citep{nie2025large,ye2025dream,bie2025llada2,bethune2026designspacetrimodalmasked} and shown that they match AR models on a range of downstream tasks while offering the prospect of substantially more efficient inference \citep{wu2025fastdllmtrainingfreeaccelerationdiffusion,ben2025accelerated,kim2025train,jazbec2025learning}.

A well-known limitation of masked dLLMs is that, once a token has been unmasked, the model has no mechanism to revise it \citep{wang2025remasking,vonrutte2026why}, even if, later in the generation process and with more context, another token would be a better fit. This stems directly from the absorbing forward process used during training; it makes masked dLLMs susceptible to error propagation and limits their overall performance. To address this, a large body of work has recently attempted to equip masked dLLMs with self-correcting capabilities, either through fine-tuning or continual pre-training \citep{kim2025fine,schiff2026learn,chen2026dmax,yu2026introspective,huang2025don}, or post-hoc (training-free) methods \citep{hong2025wide,dong2025saber,xiang2026stop}. The latter are particularly appealing as they can be applied off-the-shelf to any pretrained dLLM. A representative example is WINO \citep{hong2025wide}, which complements confidence-based adaptive unmasking (Fast-dLLM; \citealt{wu2025fastdllmtrainingfreeaccelerationdiffusion}) with a confidence-based remasking step that reverts already-unmasked tokens deemed unlikely under their current surrounding context. Closely related are Saber \citep{dong2025saber}, which remasks tokens whose confidence has dropped the most since the previous step, and COVER \citep{xiang2026stop}, which keeps, remasks, or replaces unmasked tokens based on leave-one-out confidences approximated via a KV-cache override.

All three methods report gains in accuracy and speed. These gains, however, are reported at individual operating points rather than along an accuracy--compute frontier. WINO and COVER are also compared mainly against high-confidence sampling with a fixed number of tokens per step, rather than against a tuned Fast-dLLM. WINO and Saber are moreover evaluated mainly at large block lengths ($BL \geq 128$), where confidence-based unmasking is known to degrade \citep{nie2025large}. Finally, all three are evaluated mainly under greedy decoding, although stochastic (i.e., non-greedy) decoding is central to RL post-training of dLLMs \citep{zhao2025d1,tang2025wd1} and to $\texttt{pass@}k$ evaluation. It thus remains unclear how much of the reported gains comes from remasking itself. In this work, we therefore do not propose a new remasking method. Instead, we ask when training-free, confidence-based remasking actually helps in masked dLLMs.

To this end, we first compare the accuracy--compute Pareto frontiers of WINO, Saber, and COVER with those of Fast-dLLM under greedy decoding, across two dLLMs, four benchmarks, and two block lengths (\Cref{sec:exp_takeaway1}). At the commonly used shorter block length ($BL=32$), all three perform on par with Fast-dLLM, while additionally being slower in wall-clock time. Remasking gains appear only at $BL=128$, where even WINO, the strongest of the three, still trails Fast-dLLM at $BL=32$. Analyzing WINO in depth, we find that its remasking confidences, computed via auxiliary shadow tokens, closely match oracle leave-one-out confidences. However, most remasked positions are re-unmasked to the same token, so-called \emph{flip-flops} \citep{xiang2026stop} (\Cref{sec:exp_takeaway2}).

Moving to stochastic generation (\Cref{sec:exp_takeaway34}), the picture becomes more nuanced. Under stochastic decoding, WINO mitigates some stochasticity-induced errors, improving $\texttt{pass@}1$, but aggravates the diversity collapse recently reported for confidence-based unmasking \citep{ni2026flexibility,olausson2026tale}. On top of dUltra \citep{chen2025dultra}, a learned Bernoulli unmasking policy, WINO improves accuracy by ${\sim}3.2\%$ on average at only ${\sim}2$ extra NFEs, a larger gain than on top of Fast-dLLM.

Overall, the benefits of confidence-based remasking are highly setting-dependent. Our analyses further point to the dLLM, not the remasking rule, as the bottleneck, suggesting that effective remasking may require changes to training, not only to inference.

\section{Background}
\label{sec:background}

We denote a sequence of tokens by $\vx = (x^1, \ldots, x^L) \in \mathcal{V}^L$, where $\mathcal{V}:= \{1,\ldots,V\}$ is the vocabulary and $L$ is the sequence length. Given our focus on masked discrete diffusion, the vocabulary is extended with a special mask token denoted with $\mask$. Throughout, we use the notation $[L]$ to represent the set $\{1, \ldots, L \}$ for brevity. With \emph{unmasking} we refer to $\mask \rightarrow x$ transitions and with \emph{remasking} to $x \rightarrow \mask$ (where $x \in \mathcal{V} \setminus \{\mask\} $).

\subsection{Masked diffusion language models}
Masked diffusion models are a form of absorbing discrete diffusion
\citep{austin2021, hoogeboom2021} whose noise distribution concentrates
on a single absorbing state \mask. Under the continuous-time formulation of
\citet{campbell2021}, the forward process independently replaces each token
of $\vx_0 \sim p_\text{data}$ with a mask token \mask:
\begin{align*}
p_{t|0}(\vx_t \mid \vx_0) = \prod_{k=1}^L \big( \alpha(t) \boldsymbol{1}[x_t^k = \mask] + (1 - \alpha(t)) \boldsymbol{1}[x_t^k = x_0^k] \big)
\end{align*}
where $\alpha(t)$ is a masking probability governed by a
monotonically increasing schedule $\alpha : [0,1] \to [0,1]$ with $\alpha(0)=0$ and
$\alpha(1)=1$.
The reverse (generative) process is typically modeled by training a network
$q_\theta$ to approximate the true reverse conditional $p_{0\mid t}$ directly, since its
ELBO admits the simple masked-prediction form
\citep{sahoo2024simple,shi2024simplified,ou2025}:
\begin{equation*}
\label{eq:training}
    -\mathcal{L}(\theta) \triangleq
    \mathbb{E}_{t, \vx_0, \vx_t} \left[
    \frac{1}{t}\sum_{k=1}^L \boldsymbol{1}[x_t^k = \mask]
    \log q_\theta^k(x_0^k \mid \vx_t) \right],
\end{equation*}
with $t \sim U[0,1]$, $\vx_0 \sim p_\text{data}$, $\vx_t \sim p_{t|0}$, and $q_\theta^k(\cdot \mid \vx_t)$ denoting the model's predicted marginal at position $k$.

\subsection{Unmasking in dLLMs}
\label{sec:unmasking}
At test time, the reverse process is discretized and, starting from a fully masked sequence $\vx_T := \mask^L$, new samples are obtained via progressive \emph{unmasking}. Representing the current generation as $\vx_t \in \mathcal{V}^L$ with still-masked positions $\mathcal M_t := \{k \in  [L] \mid x_t^k = \mask\}$, the next-step generation $\vx_{t-1}$ is sampled via
\begin{align*}
    x_{t-1}^k := \begin{cases}
        x \sim q_{\theta}^k(\cdot \mid \vx_t \: ; \tau), & \text{if } k \in \mathcal U_t^K, \\
        x_{t}^k\;, & \text{otherwise}
    \end{cases}
\end{align*}
where the set $\mathcal{U}_t^K$ of $K \geq 1$ unmasking positions is sampled uniformly at random (without replacement) from $\mathcal M_t$ \citep{zhengMaskedDiffusionModels2025, nie2025large} and the process is repeated until there are no masked positions left ($\mathcal{M}_t = \emptyset$). Here $\tau$ denotes a sampling temperature, with higher temperatures yielding more diverse generated samples. In dLLMs, it is common to set $\tau=0$, which corresponds to greedy token selection.

\begin{wrapfigure}{r}{0.5\linewidth}
\vspace{-24pt}
\begin{minipage}{\linewidth}
\begin{algorithm}[H]
\caption{WINO sampling algorithm \citep{hong2025wide}. Differences to Fast-dLLM \citep{wu2025fastdllmtrainingfreeaccelerationdiffusion} unmasking are denoted in \colorbox{cyan!15}{blue}.}
\label{alg:wino}
\small
\begin{algorithmic}[1]
\STATE \textbf{Input:} $\vx_t$, $b_t$, $q_\theta$, $\tau$, $\lambda_1$,  \colorbox{cyan!15}{$\lambda_2$}
\STATE \textbf{Output:} $\vx_{t-1}, b_{t-1}$
\STATEx
\STATE  \colorbox{cyan!15}{$\tilde{\vx}_t \gets [\vx_t, \mask, \ldots, \mask] \in \mathcal{V}^{L + BL}$}
\STATE $\mathcal{B} \gets [b_t \cdot BL : (b_t +1) \cdot BL]$
\STATE $\mathcal{M}_t \gets \{k \in [L] \mid x_t^k = \mask\}$
\STATEx
\STATE $q_t^k \gets q_{\theta}^k(\cdot \mid \tilde{\vx}_t)$
\STATE $x_0^k \sim  q_t^k(\cdot \:; \tau)$
\STATE $c_t^k \gets  q_t^k(x_0^k)$
\STATE $\mathcal{U}_t \gets \{k \in \mathcal{M}_t \cap \mathcal{B} \mid c_t^k > \lambda_1\}$
\STATE $\mathcal{U}_t \gets \mathcal{U}_t \cup \{\arg\max_{k \in \mathcal{M}_t \cap \mathcal{B}} c_t^k \}$
\STATEx
\STATE \colorbox{cyan!15}{$\tilde{c}_t^{k} \gets q_t^{k_s}(x_t^{k})$, $k_s = L + (k -b_t \cdot BL)$}
\STATE \colorbox{cyan!15}{$\mathcal{R}_t \gets \{ k \in \mathcal{B} \setminus \mathcal{M}_t \mid \tilde{c}_t^k < \lambda_2\}$}
\STATE \colorbox{cyan!15}{$\mathcal{R}_t \gets \texttt{Loop-guard} (\mathcal{R}_t, \mathcal{U}_{t+1})$}
\STATEx
\STATE
$x_{t-1}^k := \begin{cases} 
x_0^k \:, & k \in \mathcal{U}_t
\\
\makebox[0pt][l]{\colorbox{cyan!15}{$\mask \:,\quad k \in \mathcal{R}_t$}} & \\
x_t^k \:, & \text{else}
\end{cases}$
\STATE $b_{t-1} := \begin{cases} b_t + 1 \:, & \{k \in \mathcal{B} | x_{t-1}^k = \mask \} = \emptyset
\\
b_t \:, & \text{else}
\end{cases}$

\end{algorithmic}
\end{algorithm}
\end{minipage}
\vspace{-16pt}
\end{wrapfigure}

In practice, confidence-based unmasking is often chosen over random unmasking due to better empirical performance. Two particularly popular instantiations are \emph{high-confidence} sampling \citep{chang2022maskgit,nie2025large}, where $K$ positions with the highest token confidence\footnote{For stochastic sampling, the confidence is defined instead as $c_{t}^k \triangleq q_\theta^k(x \mid \vx_t), \: x \sim q_{\theta}^k(\cdot \mid \vx_t \: ; \tau)$.} $c_{t}^k \triangleq \max_{v} q_\theta^k(v \mid \vx_t)$ are unmasked
\begin{align}
\label{eq:high-conf}
\mathcal U_{t}^{K} &:= \underset{I
\subseteq \mathcal M_t,\ |I| = K}{\arg\max}\
\sum_{k \in I} c_{t}^k
\end{align}
and \emph{Fast-dLLM} \citep{wu2025fastdllmtrainingfreeaccelerationdiffusion}, where all positions with token confidence larger than a pre-specified threshold $\lambda_1$ are unmasked
\begin{align}
\label{eq:conf-thres}
\mathcal U_t^{\lambda} &:= \{ k \in
\mathcal M_t \mid c_t^k > \lambda_1 \} \cup \{\arg\max_{k \in \mathcal{M}_t} c_t^k \}
\end{align}
with a fallback to the position with the highest confidence to ensure progress is made in every sampling step. Due to its adaptive nature, Fast-dLLM was empirically shown to push the frontier of parallel-token generation in dLLMs \citep{wu2025fastdllmtrainingfreeaccelerationdiffusion,jazbec2025learning}.

Additionally, confidence-based samplers are often combined with semi-autoregressive (semi-AR; \citealt{arriola2025block}) sampling, which restricts unmasking to a contiguous block of $BL < L$ tokens at a time and proceeds block-by-block in a left-to-right order (i.e., once all positions in the current block are unmasked, generation moves to the next block). Empirically, semi-AR sampling has proved essential for confidence-based samplers, as using too large a block length $BL$ leads to compromised performance \citep{nie2025large}.

Recently, learned Bernoulli unmasking policies have gained traction as a scalable alternative to confidence-based heuristics like Fast-dLLM \citep{jazbec2025learning,chen2025dultra}. \emph{dUltra} \citep{chen2025dultra}, for example, jointly trains an unmasking head during RL post-training that predicts the unmasking probability of each still-masked position at every sampling step:
\begin{align*} \mathcal{U}_t &:= \{ k \in \mathcal{M}_t \mid b_t^k = 1 \}, \quad b_t^k \sim \mathrm{Ber}(s_t^k), \quad s_t^k = \sigma\!\left(f_{\phi}(\vh_{t}^{1:L})_k\right) \end{align*}
where $f_{\phi}$ is the unmasking head, $\vh_{t}^{1:L}$ denotes the dLLM's last-layer hidden states (before the unembedding matrix), and $\sigma$ is the sigmoid function. Due to the Bernoulli sampling of unmasking decisions, dUltra gives rise to stochastic generation even at $\tau = 0$.

\subsection{Confidence-based remasking in dLLMs}
In masked dLLMs, once a position has been unmasked, the model has no option to revise it, even if, at a later time step and with additional context, another token would be a better fit. This makes masked models susceptible to early sampling mistakes from which they cannot recover \citep{wang2025remasking}. Note that the inability to remask stems from the absorbing (towards $\mask$) forward process  used during training. To overcome this limitation, a growing body of literature has tried to add self-correcting capabilities to masked dLLMs (\citealt{hong2025wide,kim2025fine,huang2025don,schiff2026learn}; \emph{inter alia}). While most approaches require some form of retraining or fine-tuning, we focus here on methods that aim to achieve remasking in a post-hoc (training-free) manner.

A representative confidence-based remasking approach is \textbf{WINO} \citep{hong2025wide}, where the idea of confidence thresholding unmasking \citep{wu2025fastdllmtrainingfreeaccelerationdiffusion} is applied to remasking. To get informative remasking confidences efficiently, WINO attaches to the current generation $\vx_t$ an auxiliary block of so-called \emph{shadow} tokens $\tilde{\vx}_t = [\vx_t, \mask, \ldots, \mask] \in \mathcal{V}^{L + BL}$, one per position in the active block $b_t$. Crucially, each shadow position has the same positional embedding as its corresponding position in the original block but is prevented from attending to the (unmasked) token at the original position via a custom attention mask (see \Cref{fig:attention_masks}). This ensures that the shadow output approximates a re-prediction (starting from $\mask$ instead of the current token $x_t^k$) of that position given the surrounding context without the cost of an additional forward pass:\footnote{The approximation is exact for a single-layer transformer; for deeper models, the surrounding tokens' hidden state representations still encode information about $x_t^k$ via self-attention.}
\begin{align}
\label{eq:wino-approx}
    q_{\theta}^{k_s}(\cdot \mid \tilde{\vx}_t) \approx q_{\theta}^k(\cdot \mid \vx_{t, -k})
\end{align}
where $k_s = L + (k - b_t \cdot BL)$ is the shadow position corresponding to the $k$-th original position and  $\vx_{t, -k}$ corresponds to the current generation $\vx_t$ with the $k$-th token masked out (i.e., leave-one-out style prediction). Note that original positions do not attend to shadow positions, hence their predictive distribution is not affected by the presence of the shadow block: $q_{\theta}^{k}(\cdot \mid \tilde{\vx}_t) = q_{\theta}^k(\cdot \mid \vx_t)$ for $k \le L$.

All presently unmasked tokens in the active block whose shadow  confidence falls below a fixed threshold $\lambda_2$ are then remasked
\begin{align}
\label{eq:wino}
    \mathcal{R}_t := \{ k \in \mathcal{B} \setminus \mathcal{M}_t \mid q_{\theta}^{k_s}(x_t^k \mid \tilde{\vx}_t) < \lambda_2\}
\end{align}
where $\mathcal{B} := [b_t \cdot BL : (b_t +1) \cdot BL]$ denotes the set of indices in the active block $b_t$; see also \Cref{alg:wino}. The intuition is that if the current token $x_t^k$ would not likely be predicted under the same surrounding context (captured by the shadow predictive distribution $q_{\theta}^{k_s}$), the position is set to $\mask$ again so that the model can predict a new token for it in future sampling steps under a new context.

Note that once remasking is allowed, convergence ($\mathcal{M}_t = \emptyset$) is no longer guaranteed: sampling can enter a loop in which a given set of positions is repeatedly unmasked to the same tokens after being remasked. WINO's implementation\footnote{\url{https://github.com/Feng-Hong/WINO-DLLM/blob/main/LLaDA/decoding.py}} addresses this with a \emph{loop-guard} mechanism: whenever too many tokens get selected for remasking via confidence thresholding, i.e., $|\mathcal{R}_t| \ge |\mathcal{U}_{t+1}|$, only $K = |\mathcal{U}_{t+1}| - 1$ positions with the smallest shadow confidences are remasked, guaranteeing net progress at every step.

We additionally evaluate two other confidence-based remasking methods. \textbf{Saber} \citep{dong2025saber} tracks the dLLM's own (\emph{native}) confidence $c_t^k$ at each unmasked position and remasks the positions whose confidence dropped the most since the previous step. \textbf{COVER} \citep{xiang2026stop} approximates leave-one-out confidences $q_\theta^k(\cdot \mid \vx_{t,-k})$ by overriding the KV cache and is designed to reduce flip-flops.

\section{Re-evaluating confidence-based remasking}
\label{sec:exp}

We now ask whether confidence-based remasking improves the performance of masked dLLMs. We take the evaluation in the WINO paper \citep{hong2025wide} as a starting point, as it leaves three questions open: i) its main baseline is high-confidence sampling (Equation~\ref{eq:high-conf}), leaving it unclear whether WINO's gains come from adaptive unmasking (Fast-dLLM) or from remasking itself; ii) it evaluates mainly at large block lengths ($BL \ge 128$ instead of the standard $BL=32$), where confidence-based unmasking is known to struggle \citep{nie2025large}; and iii) it studies remasking primarily under greedy decoding ($\tau=0$), leaving open how remasking performs under stochastic decoding. We address i) and ii) in \Cref{sec:exp_takeaway1}, where we also include Saber and COVER, analyze WINO's limited gains in \Cref{sec:exp_takeaway2}, and address iii) in \Cref{sec:exp_takeaway34}.

\paragraph{Experimental setup} We evaluate on two open-source dLLMs, LLaDA-8B-Instruct \citep{nie2025large} and Dream-v0-Instruct-7B \citep{ye2025dream}, and four standard benchmarks: GSM8K \citep{cobbe2021gsm8k}, MATH-500 \citep{hendrycksmath2021}, HumanEval \citep{chen2021evaluating}, and MBPP \citep{austin2021programsynthesislargelanguage}. All generations are zero-shot, with a generation length of 256 and, unless stated otherwise, greedy decoding ($\tau=0$). We trace the Pareto frontiers of Fast-dLLM, WINO, and COVER by varying the unmasking threshold $\lambda_1 \in \{0.5, 0.6, 0.7, 0.8, 0.9\}$, and that of Saber, which sets its unmasking threshold adaptively, by varying its remasking budget $\mu \in \{2, 4, 8\}$. For WINO, we set the remasking threshold $\lambda_2=0.8$ (other values in \Cref{fig:lambda2}); for Saber and COVER, we use the official implementations with the default values of all other hyperparameters. To study the role of block length, we vary $BL \in \{32, 128\}$: $BL=32$ is the current standard setting for LLaDA-style models \citep{ni2026flexibility, wu2025fastdllmtrainingfreeaccelerationdiffusion}, while $BL=128$ is the primary setting reported in the WINO paper.

\subsection{Confidence-based remasking vs.\ unmasking alone}
\label{sec:exp_takeaway1}

We first test whether confidence-based remasking (WINO, Saber, COVER) brings meaningful gains over confidence-threshold unmasking alone (Fast-dLLM; \citealt{wu2025fastdllmtrainingfreeaccelerationdiffusion}).

\begin{figure}[h]
\centering
\includegraphics[width=\linewidth]{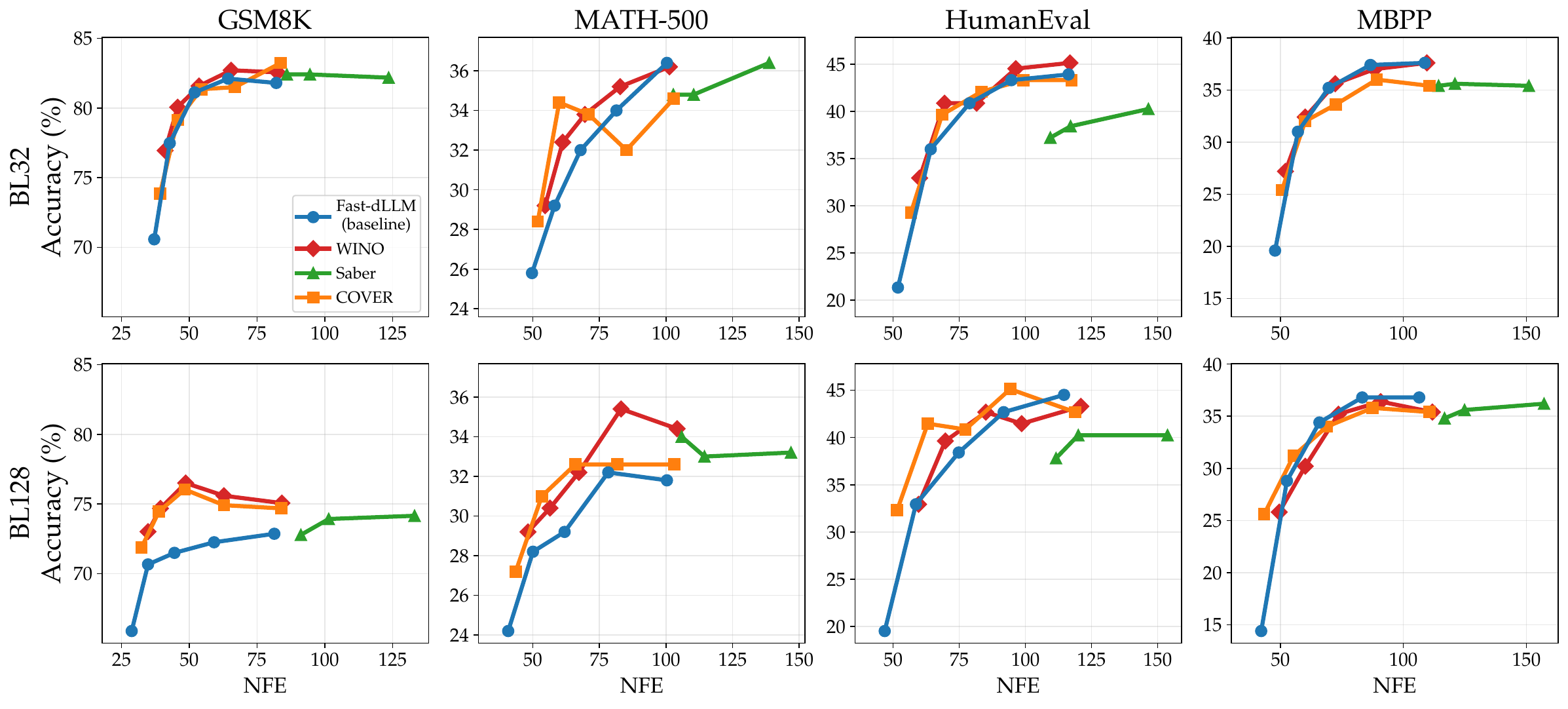}
    \caption{Pareto frontiers of accuracy vs.\ network function evaluations (NFEs) for remasking methods (WINO, Saber, COVER) and Fast-dLLM on LLaDA-8B-Instruct under greedy decoding ($\tau=0$), with block lengths $BL=32$ (top) and $BL=128$ (bottom); sweep details in \Cref{sec:exp}. At $BL=32$, no remasking method meaningfully improves over Fast-dLLM. At $BL=128$, WINO gains more than Saber and COVER, but still underperforms Fast-dLLM at $BL=32$. Results for Dream-v0-Instruct-7B are shown in \Cref{fig:wino_greedy_dream}.}
    \label{fig:wino_greedy}

\end{figure}

\begin{table}[t]
\centering
\caption{Difference in top accuracy, $\Delta$ (\%), between each remasking method and Fast-dLLM, each evaluated at its optimal point along its Pareto frontier (setup of \Cref{fig:wino_greedy,fig:wino_greedy_dream}).}
\label{tab:top_perf_delta}
\small
\setlength{\tabcolsep}{3.5pt}
\begin{tabular}{ll cccc cccc}
\toprule
 & & \multicolumn{4}{c}{$BL=32$} & \multicolumn{4}{c}{$BL=128$} \\
\cmidrule(lr){3-6} \cmidrule(lr){7-10}
Model & Method & GSM8K & MATH & HEval & MBPP & GSM8K & MATH & HEval & MBPP \\
\midrule
\multirow{3}{*}{LLaDA} & WINO & $+0.60$ & $-0.20$ & $+1.22$ & $+0.00$ & $+3.64$ & $+3.20$ & $-1.22$ & $-0.40$ \\
 & Saber & $+0.30$ & $+0.00$ & $-3.66$ & $-2.00$ & $+1.29$ & $+1.80$ & $-4.27$ & $-0.60$ \\
 & COVER & $+1.13$ & $-1.80$ & $-0.61$ & $-1.60$ & $+3.18$ & $+0.40$ & $+0.61$ & $-1.00$ \\
\midrule
\multirow{3}{*}{Dream} & WINO & $+0.38$ & $+0.60$ & $+1.22$ & $+0.00$ & $+0.45$ & $+1.60$ & $+1.83$ & $+2.00$ \\
 & Saber & $+1.29$ & $+0.00$ & $-1.22$ & $+0.00$ & $+0.68$ & $+0.20$ & $-2.44$ & $+1.00$ \\
 & COVER & $+0.68$ & $+1.80$ & $+1.22$ & $-0.60$ & $-10.39$ & $-5.80$ & $-4.88$ & $-1.60$ \\
\bottomrule
\end{tabular}
\end{table}

Starting with $BL=32$ (\Cref{fig:wino_greedy}), remasking sometimes improves performance (e.g., WINO on LLaDA's HumanEval), but the gains are generally marginal and inconsistent across datasets and along the Pareto frontier. Averaged over the four datasets, WINO improves over Fast-dLLM's top performance by ${\sim}0.4\%$ on LLaDA and ${\sim}0.5\%$ on Dream (per-dataset values in \Cref{tab:top_perf_delta}), differences that are small relative to the binomial standard errors (\Cref{fig:wino_greedy_se}). Saber changes top performance by ${\sim}{-1.3}\%$ and ${\sim}0.0\%$, and COVER by ${\sim}{-0.7}\%$ and ${\sim}0.8\%$ (\Cref{fig:wino_greedy,fig:wino_greedy_dream}). Since each WINO forward pass extends the sequence with a shadow block, a WINO step is more expensive than a Fast-dLLM step, in both FLOPs and latency. We therefore also compare wall-clock time (\Cref{fig:wallclock}): on most datasets, all three methods reach their top performance later than Fast-dLLM.

We next turn to $BL=128$, the main setting of the WINO paper \citep{hong2025wide}. Here WINO yields larger gains over Fast-dLLM (\Cref{fig:wino_greedy}), improving top performance by ${\sim}1.3\%$ on LLaDA and ${\sim}1.5\%$ on Dream on average. In wall-clock time (\Cref{fig:wallclock}), however, these gains become less consistent, as the cost of its shadow block grows with the block length: at the same unmasking threshold, WINO takes up to $1.7\times$ as long as Fast-dLLM, compared to up to $1.3\times$ at $BL=32$. On LLaDA's HumanEval and MBPP, Fast-dLLM is even ahead along most of the frontier. Saber and COVER do not consistently share these gains: Saber slightly underperforms Fast-dLLM on both models, and COVER improves by ${\sim}0.8\%$ on LLaDA but drops by ${\sim}5.7\%$ on Dream (\Cref{fig:wino_greedy_dream}); its original paper evaluates Dream only at $BL=32$.

One possible reason for WINO's gains is that a larger block length gives it more opportunity to correct mistakes over a longer range. However, masked dLLMs are known to degrade under confidence-based samplers as block length grows, since corrupted \texttt{EOS} confidences produce degenerate generation orders \citep{jazbec2025learning}. Since WINO at $BL=128$ remains below Fast-dLLM at $BL=32$, both in top accuracy and at almost every wall-clock time budget, remasking appears to repair this failure mode rather than raise the underlying dLLM's performance ceiling. Together with its marginal gains at $BL=32$, this suggests that WINO's improvements may not justify its added computational overhead and implementation complexity.

\begin{takeaway}
\textbf{Takeaway 1.} At the commonly used $BL=32$, none of the three confidence-based remasking methods meaningfully improves over Fast-dLLM, while all add wall-clock overhead. At $BL=128$, WINO mitigates the degradation of confidence-based unmasking but does not raise the performance ceiling of masked dLLMs, while Saber and COVER show no consistent gains.
\end{takeaway}

\subsection{Understanding WINO's limited gains at short block lengths}
\label{sec:exp_takeaway2}
We now ask why WINO, the strongest of the three methods, gains little at the standard $BL=32$, and whether the remasking rule or the underlying dLLM is the limiting factor.

\begin{figure}[h]
\centering
\begin{minipage}[c]{0.4\linewidth}
\centering
\includegraphics[width=\linewidth]{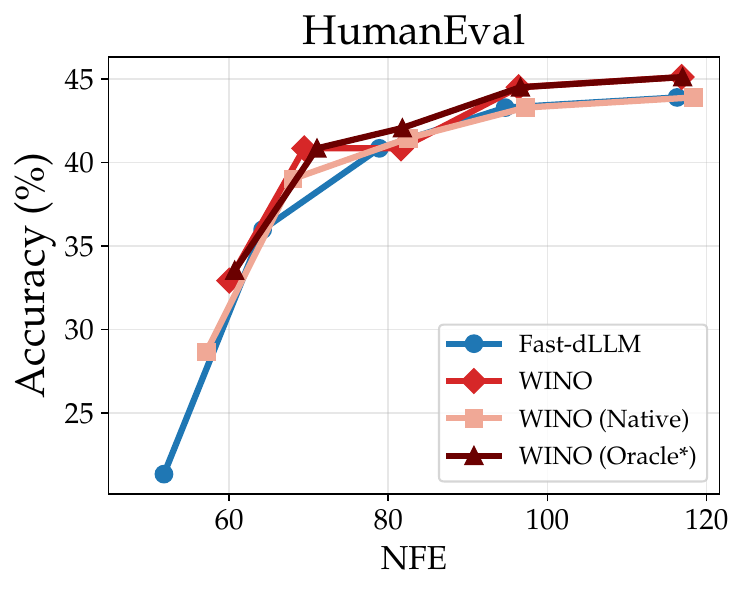}
\end{minipage}\hfill
\begin{minipage}[c]{0.57\linewidth}
\caption{Quality of WINO's shadow-token confidence as an approximation (Equation~\ref{eq:wino-approx}) to the oracle leave-one-out confidence (LLaDA-8B-Instruct, HumanEval, $BL=32$, $\lambda_2=0.8$). \emph{WINO} uses the shadow prediction $q_{\theta}^{k_s}(\cdot \mid \tilde{\vx}_t)$; \emph{WINO (Oracle*)} uses the leave-one-out prediction $q_{\theta}^k(\cdot \mid \vx_{t,-k})$, incurring one extra forward pass per candidate position; \emph{WINO (Native)} reuses the dLLM's native confidence $q_{\theta}^k(\cdot \mid \vx_t)$ at the already-unmasked position. Shadow confidences match oracle performance at a fraction of the cost, whereas the native confidence yields no gain over unmasking alone. *For the oracle, the x-axis shows sampling steps; its actual NFEs are ${\sim}14\times$ larger.}
\label{fig:faithful}
\end{minipage}
\end{figure}

\paragraph{Shadow-token approximation} We first test whether WINO's design choices limit its gains. Ablating its remasking criterion, loop-guard, shadow attention mask, and max-accept cap does not yield meaningful gains (Appendix~\ref{sec:app_ablations}), ruling out these choices as the cause. We therefore turn to its core component, the shadow-token approximation of the remasking confidence (Equation~\ref{eq:wino-approx}). Recall that, ideally, remasking would be based on the oracle leave-one-out confidence $q_{\theta}^k(\cdot \mid \vx_{t, -k})$. However, computing such oracle confidences requires an additional forward pass for every candidate remasking position at each sampling step. We nevertheless run a WINO variant that remasks based on these oracle confidences (\emph{WINO (Oracle*)}) as a reference. On LLaDA's HumanEval (\Cref{fig:faithful}), WINO's shadow confidences match the performance of \emph{WINO (Oracle*)} while being much cheaper (the oracle approach requires ${\sim}14\times$ as many NFEs). We also compare top-1 consistency between shadow and oracle predictive distributions in \Cref{fig:disagreement}, finding it to be above 96\%, further validating the shadow-token approximation. Moreover, a WINO variant that remasks based on the dLLM's native confidence $q_{\theta}^k(\cdot \mid \vx_t)$ at the unmasked position (\emph{WINO (Native)}) fails to improve over unmasking alone, confirming the necessity of shadow tokens.

\begin{figure}[h]
\centering
\includegraphics[width=\linewidth]{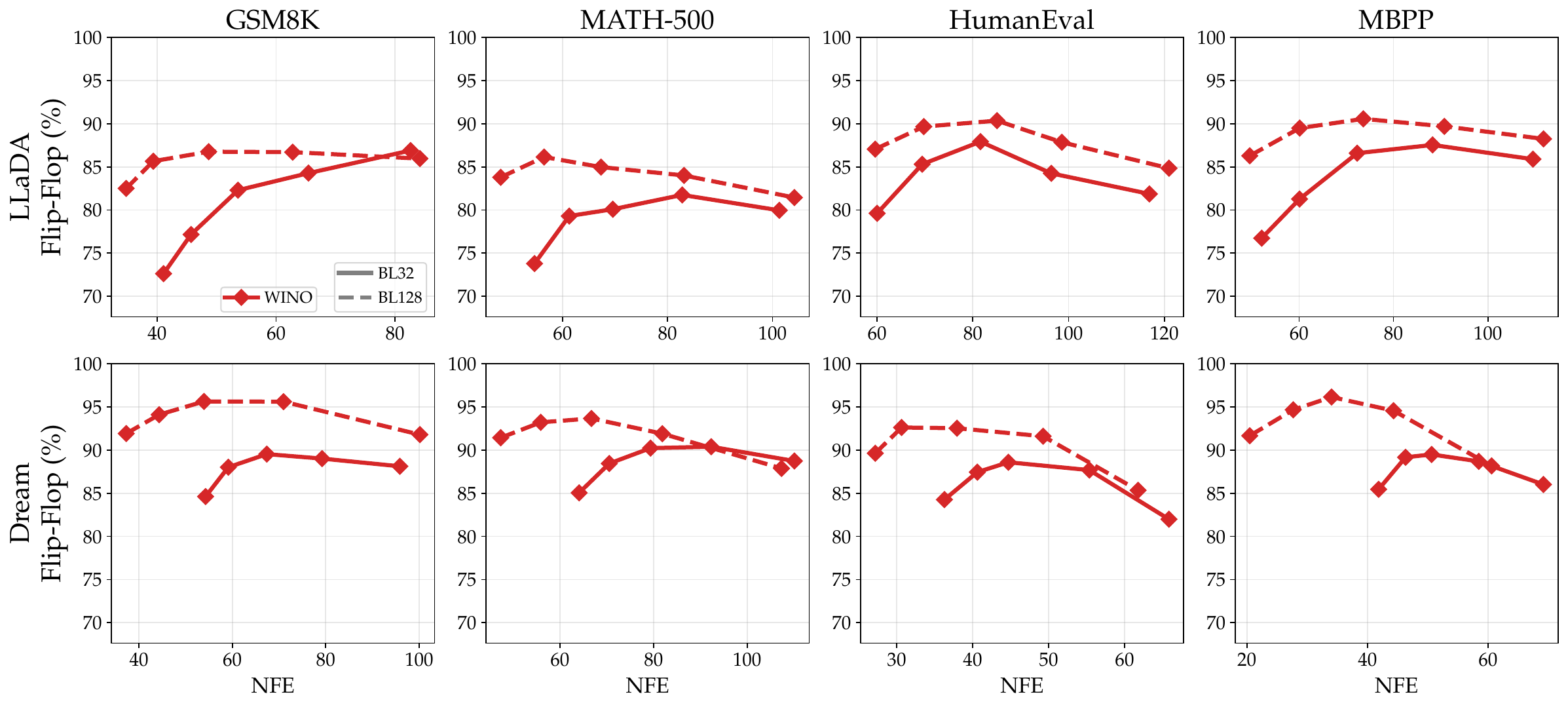}
    \caption{Flip-flop rate along the Pareto frontiers of WINO ($\lambda_2=0.8$), defined as the fraction of remasking events in which the position is re-unmasked to the same token \citep{xiang2026stop} (solid: $BL=32$, dashed: $BL=128$). ${\sim}75\%$ to ${\sim}90\%$ of remasked positions on LLaDA-8B-Instruct and ${\sim}80\%$ to ${\sim}95\%$ on Dream-v0-Instruct-7B are restored to their original token.}
    \label{fig:flipflop}
\end{figure}

\paragraph{Flip-flops} Since the shadow-token approximation does not explain WINO's limited gains, we next measure the \emph{flip-flop} rate of WINO, following \citet{xiang2026stop}. Concretely, the flip-flop rate is defined as the proportion of remasks that result in no token change, i.e., a position is remasked and then re-unmasked to the same token. As shown in \Cref{fig:flipflop}, flip-flops occur at high rates for WINO across the entire Pareto frontier: on average, ${\sim}75\%$ to ${\sim}90\%$ (LLaDA) and ${\sim}80\%$ to ${\sim}95\%$ (Dream) of remasked positions end up with the same token. Combined with the shadow-token result above, this suggests that WINO \emph{identifies} the same positions as the oracle confidence; at those positions, however, the dLLM rarely \emph{proposes} a different token. Moreover, flip-flop rates are nearly identical for generations with correct and incorrect final answers (Appendix~\ref{sec:app_ff_correctness}), showing that flip-flops are not specific to failed generations. 

\paragraph{Cascading dependencies} One possible explanation for high flip-flop rates is that tokens unmasked later were conditioned on $x_t^k$ and still encode it after position $k$ is remasked, pinning the prediction back to $x_t^k$. To test this, we additionally remask the temporal or spatial neighbors of each flagged position, i.e., tokens unmasked at the same or the next sampling step, or its left and right neighbors (Appendix~\ref{sec:app_expand_remask}). Surprisingly, both variants match or degrade accuracy relative to remasking only the flagged position. They also do not reduce the flip-flop rate, and often increase it. Sampling remasked positions stochastically ($\tau=1$) instead of greedily also does not help, as it only slightly lowers flip-flop rates and does not improve accuracy (Appendix~\ref{sec:app_stoch_remask}). This indicates that the bottleneck is neither cascading dependencies nor greedy re-prediction, but the dLLM's predictive distribution at position $k$ itself, which strongly favors the original token.
\begin{takeaway}
\textbf{Takeaway 2.} WINO's shadow-token confidence closely approximates the oracle leave-one-out remasking confidence. At the flagged positions, however, the dLLM mostly re-proposes the original token, even when neighboring tokens are also remasked or the token is sampled stochastically. This points to the dLLM, rather than the remasking rule, as the bottleneck.
\end{takeaway}

\subsection{Remasking under stochastic generation}
\label{sec:exp_takeaway34}

Lastly, we study whether confidence-based remasking is beneficial under stochastic generation. Although this setting is largely absent from the current literature on remasking in dLLMs, we believe it is particularly relevant for remasking approaches, as the increased stochasticity may produce more erroneous tokens that could benefit from revision. We consider two variants of stochastic generation: i) stochastic decoding ($\tau > 0$) with deterministic unmasking (Fast-dLLM) and ii) greedy decoding with stochastic unmasking (via learned Bernoulli policies; \citealt{jazbec2025learning,chen2025dultra}).

\begin{figure}[h]
\centering
\includegraphics[width=\linewidth]{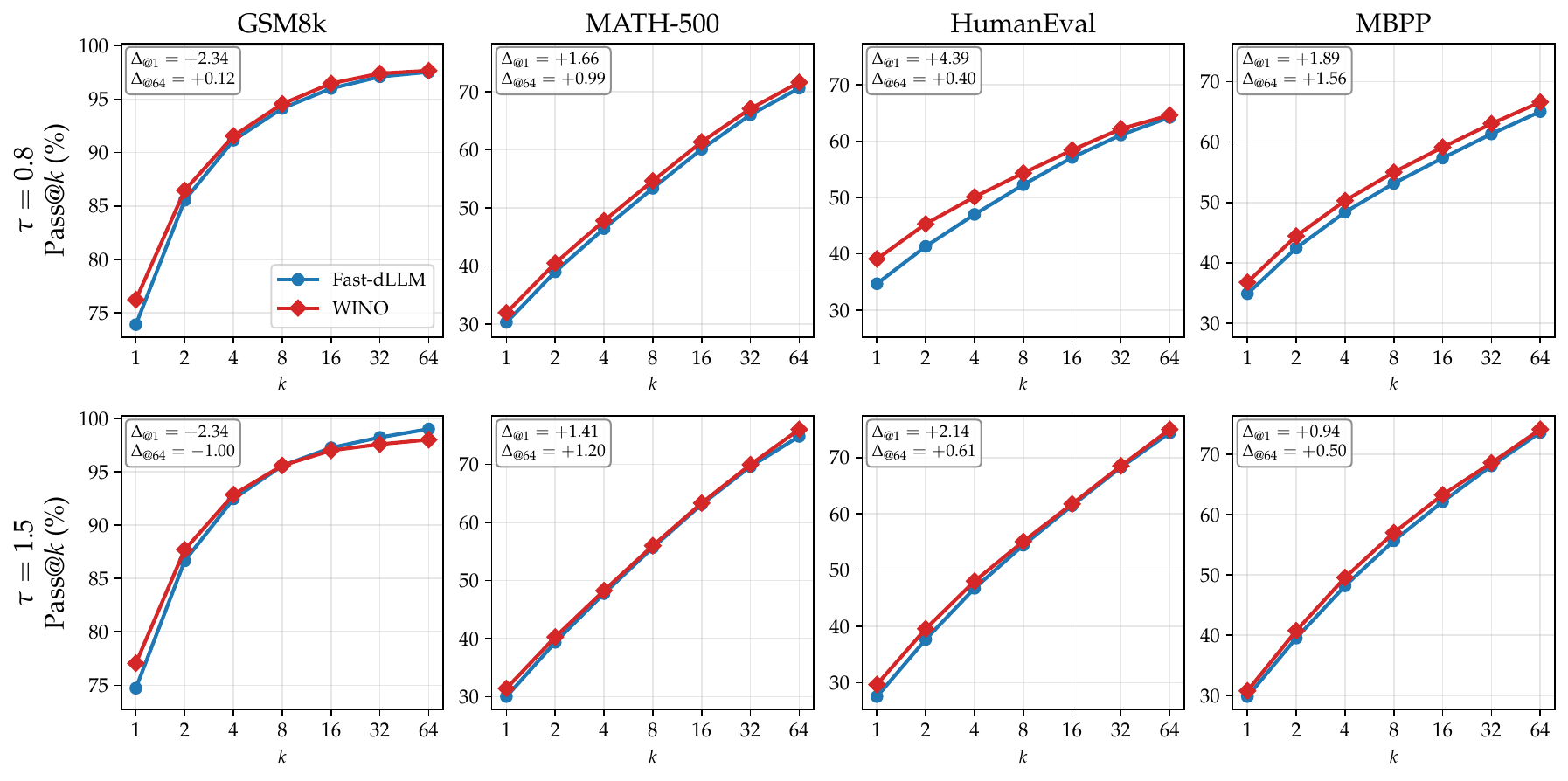}
    \caption{$\texttt{pass@}k$ for WINO and Fast-dLLM under stochastic decoding (LLaDA-8B-Instruct, $BL=32$, $\lambda_1=0.6$, $\lambda_2=0.8$) at temperatures
 $\tau\in\{0.8,1.5\}$. WINO improves $\texttt{pass@}1$ by ${\sim}2.6\%$ on average at $\tau=0.8$, but its advantage diminishes as $k$ grows, shrinking to ${\sim}0.7\%$ on average at $\texttt{pass@}64$. This suggests that remasking further constrains generation diversity on top of the diversity collapse already documented for confidence-based unmasking \citep{ni2026flexibility}.}
    \label{fig:wino_passatk}
\end{figure}

\paragraph{$\texttt{pass@}k$ scaling} We compare Fast-dLLM ($\lambda_1 = 0.6$) and WINO ($\lambda_1 = 0.6$) on LLaDA-8B-Instruct with $BL=32$ under stochastic decoding. Concretely, we consider $\tau=0.8$, a standard temperature in RL post-training of dLLMs \citep{zhao2025d1,tang2025wd1}, as well as $\tau=1.5$, motivated by the observation that dLLMs benefit from higher sampling temperatures than AR LLMs \citep{ni2026flexibility}. To measure performance, we report the commonly used $\texttt{pass@}k$ metric\footnote{Following \citet{chen2021evaluating}, we report the unbiased estimator $\texttt{pass@}k := \mathbb{E}\!\left[1 - \binom{n-c}{k}\big/\binom{n}{k}\right]$, where $n \geq k$ is the number of samples generated per problem and $c$ is the number of those that are correct.}, which tracks the probability that at least one of $k$ generated solutions is correct. As shown in \Cref{fig:wino_passatk}, remasking improves $\texttt{pass@}1$ by ${\sim}2.6\%$ on average across the four datasets at $\tau = 0.8$. However, remasking also appears to hurt generation diversity: WINO scales worse than Fast-dLLM as $k$ grows, with gains shrinking to ${\sim}0.7\%$ at $\texttt{pass@}64$ for $\tau = 0.8$. This suggests that confidence-based remasking may further exacerbate the diversity collapse previously observed for confidence-based unmasking \citep{ni2026flexibility, olausson2026tale}, where greedily relying on confidence to decide which positions to sample constrains exploration of the solution space. Consistent with this, WINO lowers the mean token entropy compared to Fast-dLLM ($0.61$ vs.\ $0.64$ on MATH-500), in part because remasked positions are re-committed with much lower entropy ($1.15 \rightarrow 0.43$; \Cref{tab:entropy}). At $\tau=1.5$, WINO's improvements shrink further, to ${\sim}1.6\%$ and ${\sim}0.3\%$ for $\texttt{pass@}1$ and $\texttt{pass@}64$, respectively.

\begin{takeaway}
\textbf{Takeaway 3.} Under stochastic decoding, WINO improves $\texttt{pass@}1$ but exacerbates the diversity collapse of confidence-based unmasking, with gains shrinking as the number of samples $k$ grows.
\end{takeaway}

\begin{figure}[h]
\centering
\includegraphics[width=\linewidth]{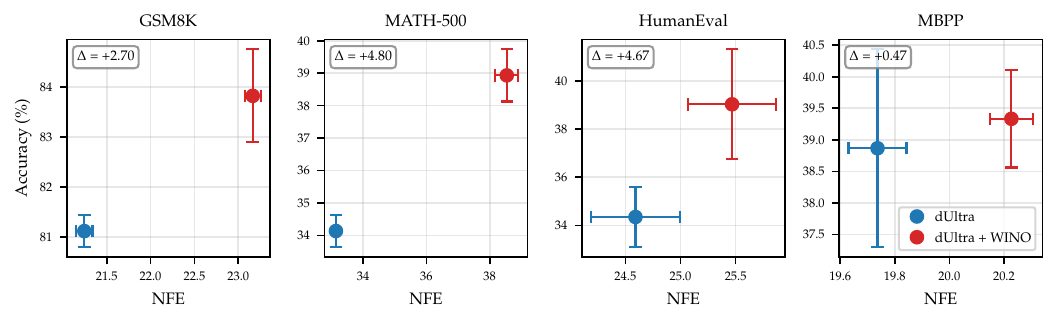}
    \caption{Accuracy and NFEs for dUltra with and without WINO remasking on LLaDA-8B-Instruct ($BL=32$, $\lambda_2=0.8$, $\tau=0$); means and error bars are over 3 seeds. Adding WINO remasking on top of dUltra's Bernoulli unmasker yields an average accuracy gain of ${\sim}3.2\%$ across the four benchmarks at a cost of only ${\sim}2$ extra NFEs, suggesting that the value of remasking depends on the unmasking strategy.}
    \label{fig:dultra}
\end{figure}

\paragraph{Bernoulli unmasking} We next keep greedy token selection ($\tau = 0$) but make the unmasking decisions stochastic, using the learned Bernoulli unmasking policy of dUltra (\Cref{sec:unmasking}). Results comparing dUltra with and without WINO remasking at $BL=32$ are shown in \Cref{fig:dultra}; we report average accuracy and NFEs on test questions over 3 random seeds (with error bars). Adding WINO remasking on top of dUltra improves accuracy by ${\sim}3.2\%$ on average across datasets, while increasing NFEs by only ${\sim}2$ on average. This highlights the potential of confidence-based remasking when paired with stochastic unmasking strategies: it appears effective at correcting low-quality tokens introduced by the randomness of Bernoulli unmasking. Moreover, dUltra produces faster generations than Fast-dLLM (e.g., on GSM8K, it reaches ${\sim}80\%$ accuracy in only ${\sim}20$ NFEs, versus ${\sim}50$ for Fast-dLLM; see \Cref{fig:wino_greedy}), which may further explain WINO's stronger gains here: more aggressive generation yields more tokens in need of revision.

\begin{takeaway}
\textbf{Takeaway 4.} When paired with stochastic Bernoulli unmasking, WINO remasking is more effective than with deterministic Fast-dLLM unmasking, suggesting that the added value of confidence-based remasking depends on the unmasking strategy.
\end{takeaway}

\section{Related work}

\paragraph{Remasking in dLLMs} Several recent works explore remasking in masked dLLMs. One direction focuses on train-time interventions: \citet{kim2025fine} fine-tune dLLMs for provable self-correction, \citet{schiff2026learn} train models to identify and revise their own errors, \citet{liu2026teach} train a separate correction head on top of a frozen dLLM, \citet{bie2026llada21} jointly train mask-to-token and token-to-token editing objectives at scale, \citet{yu2026introspective} train dLLMs to verify their own previously generated tokens, and \citet{huang2025don} use post-training to learn when to remask. A related direction is to move away from purely absorbing (masked) diffusion to a uniform one that supports remasking `naturally': \citet{vonruette2025gidd} train hybrid mask+uniform diffusion models, \citet{wang2026generalized} simplify this with purely uniform transitions, and \citet{chen2026dmax} extend pretrained masked dLLMs into self-correcting uniform diffusion models. Our work focuses on a complementary, training-free direction.

\paragraph{Post-hoc remasking} Several recent methods enable remasking without
retraining, typically by applying a confidence-based criterion at inference
time. WINO \citep{hong2025wide}, the focus of our re-evaluation, computes
remasking confidences via auxiliary shadow tokens that approximate a
leave-one-out re-prediction without an additional forward pass. ReMDM \citep{wang2025remasking} introduces a remasking sampler that can be applied to pretrained masked diffusion models to enable revisions during generation. Saber \citep{dong2025saber} remasks the unmasked positions whose confidence dropped the most since the previous step. COVER \citep{xiang2026stop} documents the high \emph{flip-flop} rates of
confidence-based remasking and reduces them by approximating leave-one-out confidences via a KV-cache override. While these works each report gains in accuracy or speed over static confidence-based unmasking (Equation~\ref{eq:high-conf}), how much these gains depend on the evaluation setting, such as block length, greedy vs.\ stochastic decoding, and unmasking strategy, has not been systematically examined, a gap our work aims to fill.

\section{Conclusion}
\label{sec:conclusion}

To better understand the added value of confidence-based remasking in masked dLLMs, we re-evaluated three recently proposed training-free methods: WINO, Saber, and COVER. Our findings suggest that their benefits depend on the setting. Under greedy decoding at the standard $BL=32$, all three perform on par with unmasking alone (Fast-dLLM), especially once their wall-clock overhead is accounted for; gains appear only at $BL=128$, where remasking mitigates the degradation of confidence-based unmasking. Under stochastic decoding, WINO improves $\texttt{pass@}1$ at the cost of generation diversity, and it yields larger gains when paired with learned Bernoulli unmasking. Our analyses further point to the dLLM, not the remasking rule, as the bottleneck. Since the impact of remasking is highly setting-dependent, we hope that future work adopts the evaluation framework outlined here to better understand the limits and potential of confidence-based remasking in masked dLLMs.

\paragraph{Limitations and future work} While we evaluate three confidence-based remasking methods (WINO, Saber, COVER) on two dLLMs, our failure-mode analyses and the stochastic decoding experiments focus on WINO; future work should apply these analyses to other post-hoc remasking methods, including ReMDM \citep{wang2025remasking}. Our results suggest that effective remasking in masked dLLMs likely requires either some form of retraining or a shift to uniform discrete diffusion \citep{vonrutte2026why,chen2026dmax}, possibly because masked diffusion training induces a strong ``unmask-only'' prior ($\mask \rightarrow x$) that post-hoc remasking cannot easily override. Moreover, while we deliberately focused on post-hoc remasking, it would be worthwhile to explore whether the negative impact of remasking on sampling diversity (\Cref{fig:wino_passatk}) also holds for approaches that incorporate remasking via retraining or fine-tuning \citep{schiff2026learn,chen2026dmax}.

\paragraph{Impact statement} This paper presents work whose goal is to advance the field of Machine Learning. There are many potential societal consequences of our work, none which we feel must be specifically highlighted here.

\begin{ack}
We thank Theo X. Olausson, the Apple MLR Paris team, Guoxuan Xia, and Izabela Kurek for helpful discussions. This project was generously supported by the Bosch Center for Artificial Intelligence. Eric Nalisnick did not utilize resources from Johns Hopkins University for this project.
\end{ack}

\bibliography{main}
\bibliographystyle{plainnat}

\newpage
\appendix

\section{Additional figures}
\label{app:add_figs}

\begin{figure}[!h]
  \centering
  \begin{subfigure}[h]{0.45\textwidth}
    \centering
    \resizebox{\linewidth}{!}{
  \begin{tikzpicture}[scale=0.3]

    \definecolor{attended}{RGB}{144,238,144}
    \definecolor{masked}{RGB}{220,220,220}

    \newcommand{\fillcell}[3]{%
      \fill[#3] (#1, #2) rectangle (#1+1, #2+1);
      \draw[black, thin] (#1, #2) rectangle (#1+1, #2+1);
    }

    \foreach \col in {0,1,2}          { \foreach \row in {0,1,2} { \fillcell{\col}{-\row}{attended} } }
    \foreach \col in {3.5,4.5,5.5}    { \foreach \row in {0,1,2} { \fillcell{\col}{-\row}{attended} } }
    \foreach \col in {7,8,9}          { \foreach \row in {0,1,2} { \fillcell{\col}{-\row}{attended} } }
    \foreach \col in {10.5,11.5,12.5} { \foreach \row in {0,1,2} { \fillcell{\col}{-\row}{masked}   } }

    \foreach \col in {0,1,2}          { \foreach \row in {0,1,2} { \fillcell{\col}{-3.5-\row}{attended} } }
    \foreach \col in {3.5,4.5,5.5}    { \foreach \row in {0,1,2} { \fillcell{\col}{-3.5-\row}{attended} } }
    \foreach \col in {7,8,9}          { \foreach \row in {0,1,2} { \fillcell{\col}{-3.5-\row}{attended} } }
    \foreach \col in {10.5,11.5,12.5} { \fillcell{\col}{-3.5}{masked} }
    \foreach \col in {10.5,11.5,12.5} { \fillcell{\col}{-4.5}{masked} }
    \foreach \col in {10.5,11.5,12.5} { \fillcell{\col}{-5.5}{masked} }

    \foreach \col in {0,1,2}          { \foreach \row in {0,1,2} { \fillcell{\col}{-7-\row}{attended} } }
    \foreach \col in {3.5,4.5,5.5}    { \foreach \row in {0,1,2} { \fillcell{\col}{-7-\row}{attended} } }
    \foreach \col in {7,8,9}          { \foreach \row in {0,1,2} { \fillcell{\col}{-7-\row}{attended} } }
    \foreach \col in {10.5,11.5,12.5} { \foreach \row in {0,1,2} { \fillcell{\col}{-7-\row}{masked}   } }

    \foreach \col in {0,1,2} { \foreach \row in {0,1,2} { \fillcell{\col}{-10.5-\row}{attended} } }
    \fillcell{3.5}{-10.5}{masked}   \fillcell{4.5}{-10.5}{attended} \fillcell{5.5}{-10.5}{attended}
    \fillcell{3.5}{-11.5}{attended} \fillcell{4.5}{-11.5}{masked}   \fillcell{5.5}{-11.5}{attended}
    \fillcell{3.5}{-12.5}{attended} \fillcell{4.5}{-12.5}{attended} \fillcell{5.5}{-12.5}{masked}
    \foreach \col in {7,8,9}          { \foreach \row in {0,1,2} { \fillcell{\col}{-10.5-\row}{attended} } }
    \foreach \col in {10.5,11.5,12.5} { \foreach \row in {0,1,2} { \fillcell{\col}{-10.5-\row}{attended} } }

    \foreach \i/\x in {0/0, 1/1, 2/2}              { \node[above, font=\small] at (\x+0.5, 1) {\i}; }
    \foreach \i/\x in {3/3.5, 4/4.5, 5/5.5}        { \node[above, font=\small] at (\x+0.5, 1) {\i}; }
    \foreach \i/\x in {6/7, 7/8, 8/9}              { \node[above, font=\small] at (\x+0.5, 1) {\i}; }
    \foreach \i/\x in {3/10.5, 4/11.5, 5/12.5}     { \node[above, font=\small] at (\x+0.5, 1) {\i}; }

    \draw[decorate, decoration={brace, amplitude=4pt, mirror}]
      (0,-13) -- (3,-13)
      node[midway, below=5pt, font=\small] {$Y_{\text{left}}$};

    \draw[decorate, decoration={brace, amplitude=4pt, mirror}]
      (3.5,-13) -- (6.5,-13)
      node[midway, below=5pt, font=\small] {$Y_{\text{curr}}$};

    \draw[decorate, decoration={brace, amplitude=4pt, mirror}]
      (7,-13) -- (10,-13)
      node[midway, below=5pt, font=\small] {$Y_{\text{right}}$};

    \draw[decorate, decoration={brace, amplitude=4pt, mirror}]
      (10.5,-13) -- (13.5,-13)
      node[midway, below=5pt, font=\small] {$Y_{\text{shad}}$};

    \node[left, font=\small\bfseries] at (0, 1.75) {\textbf{Pos ID}};
    \foreach \i in {0,1,2} { \node[left, font=\small] at (-0.5, -\i+0.5)      {\i}; }
    \foreach \i in {0,1,2} { \node[left, font=\small] at (-0.5, -3.5-\i+0.5)  {\pgfmathparse{int(\i+3)}\pgfmathresult}; }
    \foreach \i in {0,1,2} { \node[left, font=\small] at (-0.5, -7-\i+0.5)    {\pgfmathparse{int(\i+6)}\pgfmathresult}; }
    \foreach \i in {0,1,2} { \node[left, font=\small] at (-0.5, -10.5-\i+0.5) {\pgfmathparse{int(\i+3)}\pgfmathresult}; }

  \end{tikzpicture}}
    \caption{WINO attention mask and position IDs for LLaDA-8B \citep{nie2025large}.}
    \label{fig:attention_mask_wino}
  \end{subfigure}
  \hfill
  \begin{subfigure}[h]{0.45\textwidth}
    \centering
    \resizebox{\linewidth}{!}{
  \begin{tikzpicture}[scale=0.3]

    \definecolor{attended}{RGB}{144,238,144}
    \definecolor{masked}{RGB}{220,220,220}

    \newcommand{\fillcell}[3]{%
      \fill[#3] (#1, #2) rectangle (#1+1, #2+1);
      \draw[black, thin] (#1, #2) rectangle (#1+1, #2+1);
    }

    \foreach \col in {0,1,2}          { \foreach \row in {0,1,2} { \fillcell{\col}{-\row}{attended} } }
    \foreach \col in {3.5,4.5,5.5}    { \foreach \row in {0,1,2} { \fillcell{\col}{-\row}{attended} } }
    \foreach \col in {7,8,9}          { \foreach \row in {0,1,2} { \fillcell{\col}{-\row}{attended} } }
    \foreach \col in {10.5,11.5,12.5} { \foreach \row in {0,1,2} { \fillcell{\col}{-\row}{masked}   } }

    \foreach \col in {0,1,2}          { \foreach \row in {0,1,2} { \fillcell{\col}{-3.5-\row}{attended} } }
    \foreach \col in {3.5,4.5,5.5}    { \foreach \row in {0,1,2} { \fillcell{\col}{-3.5-\row}{attended} } }
    \foreach \col in {7,8,9}          { \foreach \row in {0,1,2} { \fillcell{\col}{-3.5-\row}{attended} } }
    \foreach \col in {10.5,11.5,12.5} { \fillcell{\col}{-3.5}{masked} }
    \foreach \col in {10.5,11.5,12.5} { \fillcell{\col}{-4.5}{masked} }
    \foreach \col in {10.5,11.5,12.5} { \fillcell{\col}{-5.5}{masked} }

    \foreach \col in {0,1,2}          { \foreach \row in {0,1,2} { \fillcell{\col}{-7-\row}{attended} } }
    \foreach \col in {3.5,4.5,5.5}    { \foreach \row in {0,1,2} { \fillcell{\col}{-7-\row}{attended} } }
    \foreach \col in {7,8,9}          { \foreach \row in {0,1,2} { \fillcell{\col}{-7-\row}{attended} } }
    \foreach \col in {10.5,11.5,12.5} { \foreach \row in {0,1,2} { \fillcell{\col}{-7-\row}{masked}   } }

    \foreach \col in {0,1,2} { \foreach \row in {0,1,2} { \fillcell{\col}{-10.5-\row}{attended} } }
    \fillcell{2}{-10.5}{masked}
    \fillcell{3.5}{-10.5}{masked}   \fillcell{4.5}{-10.5}{attended} \fillcell{5.5}{-10.5}{attended}
    \fillcell{3.5}{-11.5}{masked} \fillcell{4.5}{-11.5}{masked}   \fillcell{5.5}{-11.5}{attended}
    \fillcell{3.5}{-12.5}{attended} \fillcell{4.5}{-12.5}{masked} \fillcell{5.5}{-12.5}{masked}
    \foreach \col in {7,8,9}          { \foreach \row in {0,1,2} { \fillcell{\col}{-10.5-\row}{attended} } }
    \foreach \col in {10.5,11.5,12.5} { \foreach \row in {0,1,2} { \fillcell{\col}{-10.5-\row}{attended} } }

    \foreach \i/\x in {0/0, 1/1, 2/2}              { \node[above, font=\small] at (\x+0.5, 1) {\i}; }
    \foreach \i/\x in {3/3.5, 4/4.5, 5/5.5}        { \node[above, font=\small] at (\x+0.5, 1) {\i}; }
    \foreach \i/\x in {6/7, 7/8, 8/9}              { \node[above, font=\small] at (\x+0.5, 1) {\i}; }
    \foreach \i/\x in {2/10.5, 3/11.5, 4/12.5}     { \node[above, font=\small\bfseries] at (\x+0.5, 1) {\i}; }

    \draw[decorate, decoration={brace, amplitude=4pt, mirror}]
      (0,-13) -- (3,-13)
      node[midway, below=5pt, font=\small] {$Y_{\text{left}}$};

    \draw[decorate, decoration={brace, amplitude=4pt, mirror}]
      (3.5,-13) -- (6.5,-13)
      node[midway, below=5pt, font=\small] {$Y_{\text{curr}}$};

    \draw[decorate, decoration={brace, amplitude=4pt, mirror}]
      (7,-13) -- (10,-13)
      node[midway, below=5pt, font=\small] {$Y_{\text{right}}$};

    \draw[decorate, decoration={brace, amplitude=4pt, mirror}]
      (10.5,-13) -- (13.5,-13)
      node[midway, below=5pt, font=\small] {$Y_{\text{shad}}$};

    \node[left, font=\small\bfseries] at (0, 1.75) {\textbf{Pos ID}};
    \foreach \i in {0,1,2} { \node[left, font=\small] at (-0.5, -\i+0.5)      {\i}; }
    \foreach \i in {0,1,2} { \node[left, font=\small] at (-0.5, -3.5-\i+0.5)  {\pgfmathparse{int(\i+3)}\pgfmathresult}; }
    \foreach \i in {0,1,2} { \node[left, font=\small] at (-0.5, -7-\i+0.5)    {\pgfmathparse{int(\i+6)}\pgfmathresult}; }
    \foreach \i in {0,1,2} { \node[left, font=\small\bfseries] at (-0.5, -10.5-\i+0.5) {\pgfmathparse{int(\i+2)}\pgfmathresult}; }

  \end{tikzpicture}}
    \caption{Adaptation for Dream-7B model \citep{ye2025dream}.}
    \label{fig:attention_mask_dream}
  \end{subfigure}
  \caption{WINO attention masks and position IDs: LLaDA (a) vs.\ our adaptation for Dream (b). Note that adaptation is needed since Dream was converted into a dLLM starting from an AR model. Bold indices in (b) highlight the modified position IDs of $Y_{\text{shad}}$. The shadow position IDs were shifted to obtain predictions for the current block, while the masking of main tokens is sub-diagonal and diagonal to prevent information leak due to Dream's shifted predictions.}
  \label{fig:attention_masks}
\end{figure}

\begin{figure}[!htbp]
\vspace{1.5em}
\centering
\includegraphics[width=\linewidth]{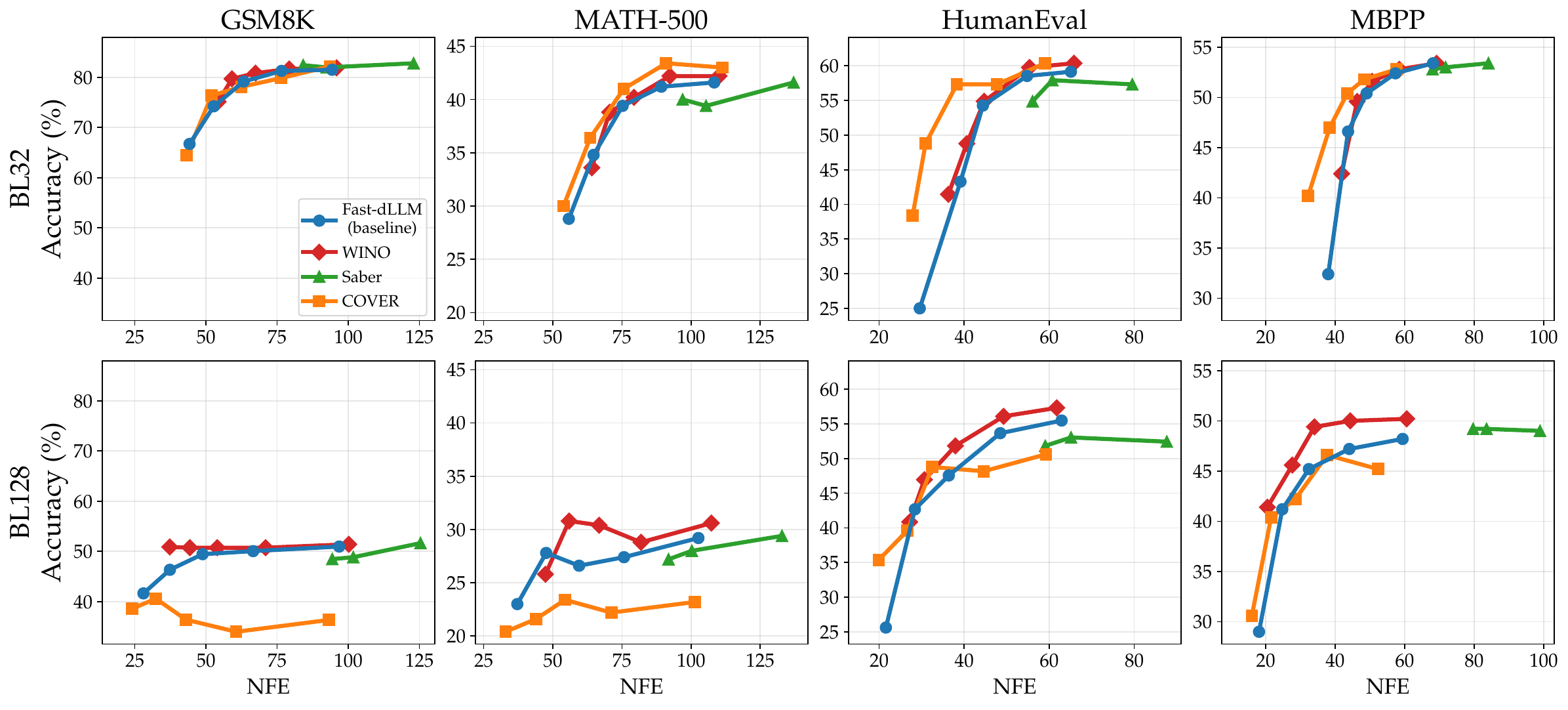}
    \caption{\Cref{fig:wino_greedy} replicated for Dream-v0-Instruct-7B.}
    \label{fig:wino_greedy_dream}
\end{figure}

\begin{figure}[!htbp]
\centering
\begin{subfigure}{\linewidth}
    \centering
    \includegraphics[width=\linewidth]{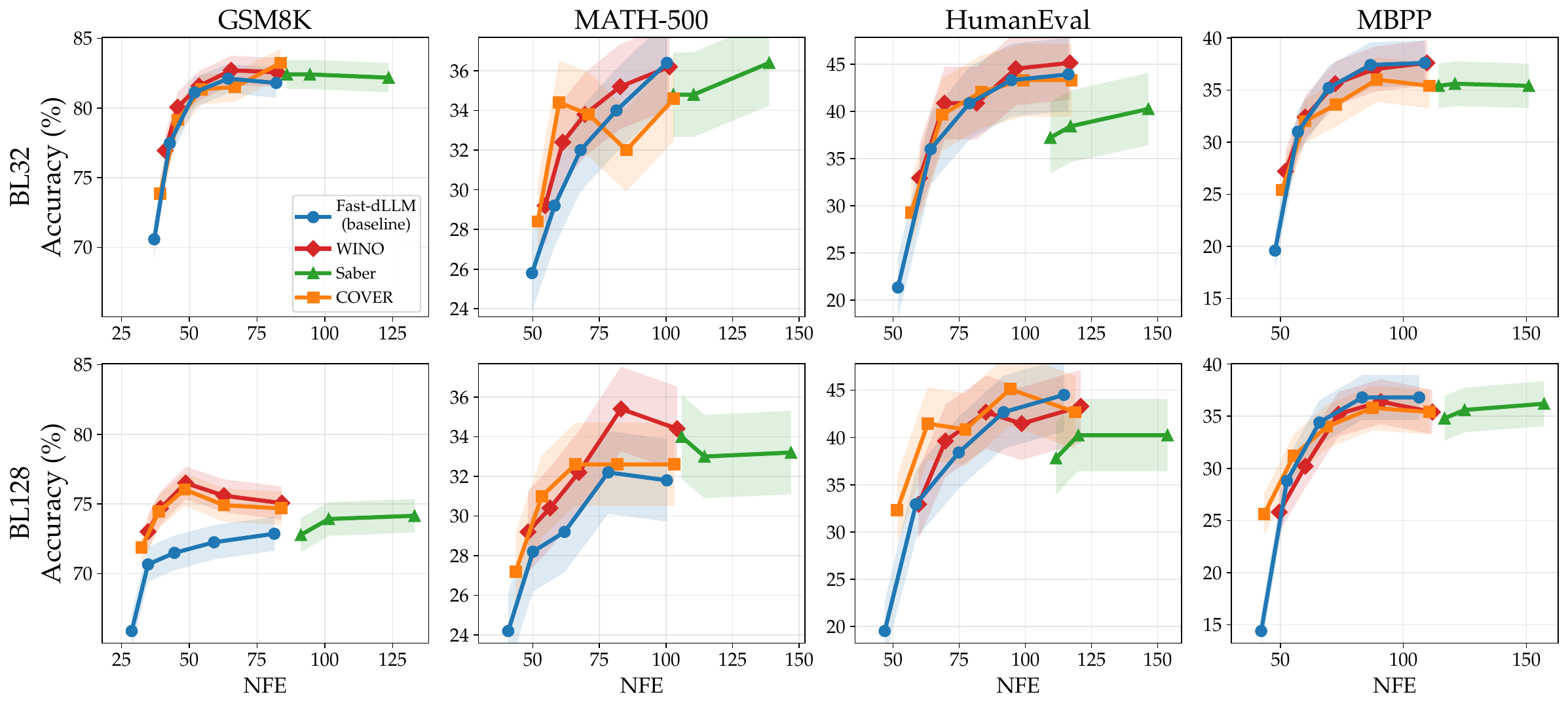}
    \caption{LLaDA-8B-Instruct}
\end{subfigure}

\vspace{0.5em}
\begin{subfigure}{\linewidth}
    \centering
    \includegraphics[width=\linewidth]{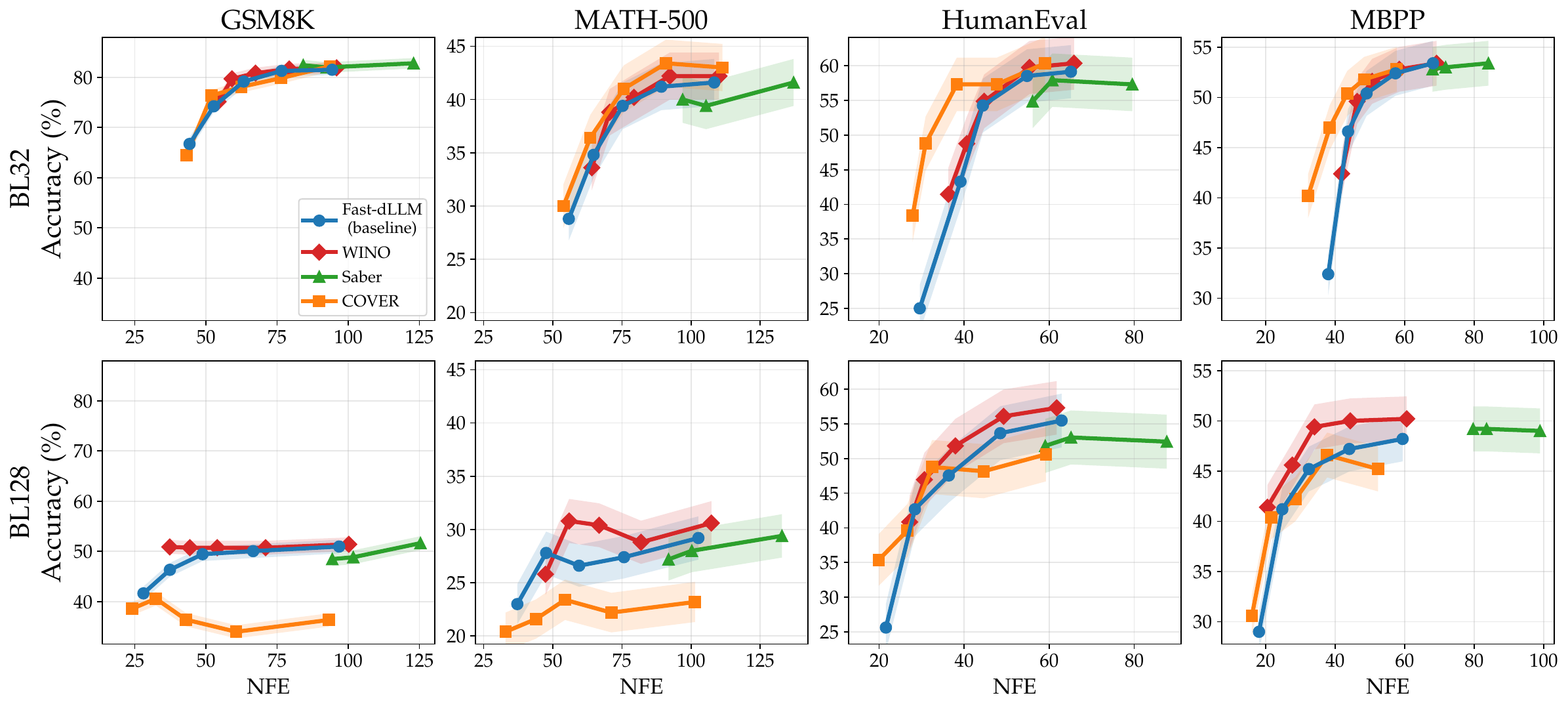}
    \caption{Dream-v0-Instruct-7B}
\end{subfigure}
    \caption{\Cref{fig:wino_greedy,fig:wino_greedy_dream} with shaded bands showing $\pm 1$ binomial standard error of the test-set accuracy, $\sqrt{\hat{p}(1-\hat{p})/n}$, where $\hat{p}$ is the accuracy and $n$ the number of test examples. At $BL=32$, WINO's average improvements over Fast-dLLM's top performance (${\sim}0.4\%$ on LLaDA and ${\sim}0.5\%$ on Dream) are small relative to these standard errors.}
    \label{fig:wino_greedy_se}
\end{figure}

\begin{figure}[!htbp]
\centering
\begin{subfigure}{\linewidth}
    \centering
    \includegraphics[width=\linewidth]{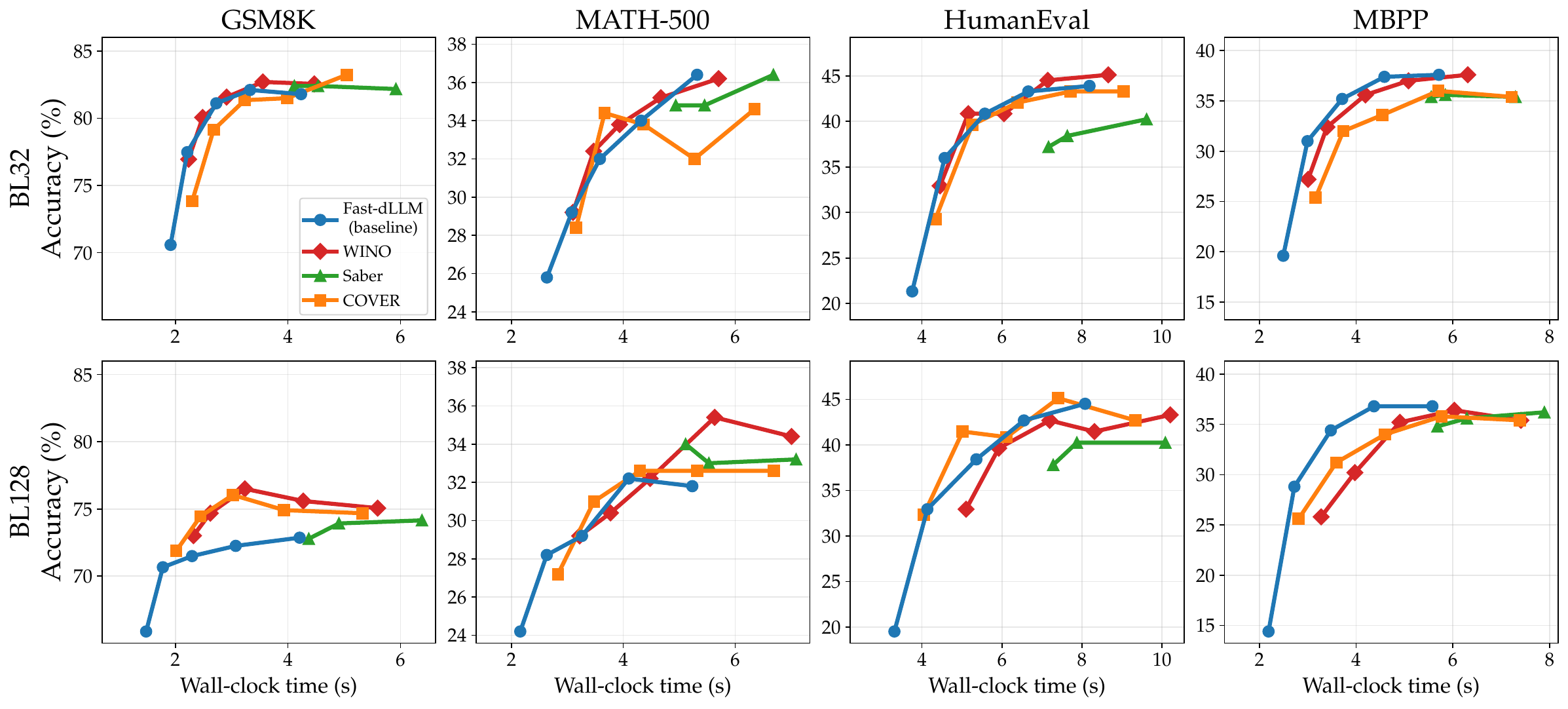}
    \caption{LLaDA-8B-Instruct}
\end{subfigure}

\vspace{0.5em}
\begin{subfigure}{\linewidth}
    \centering
    \includegraphics[width=\linewidth]{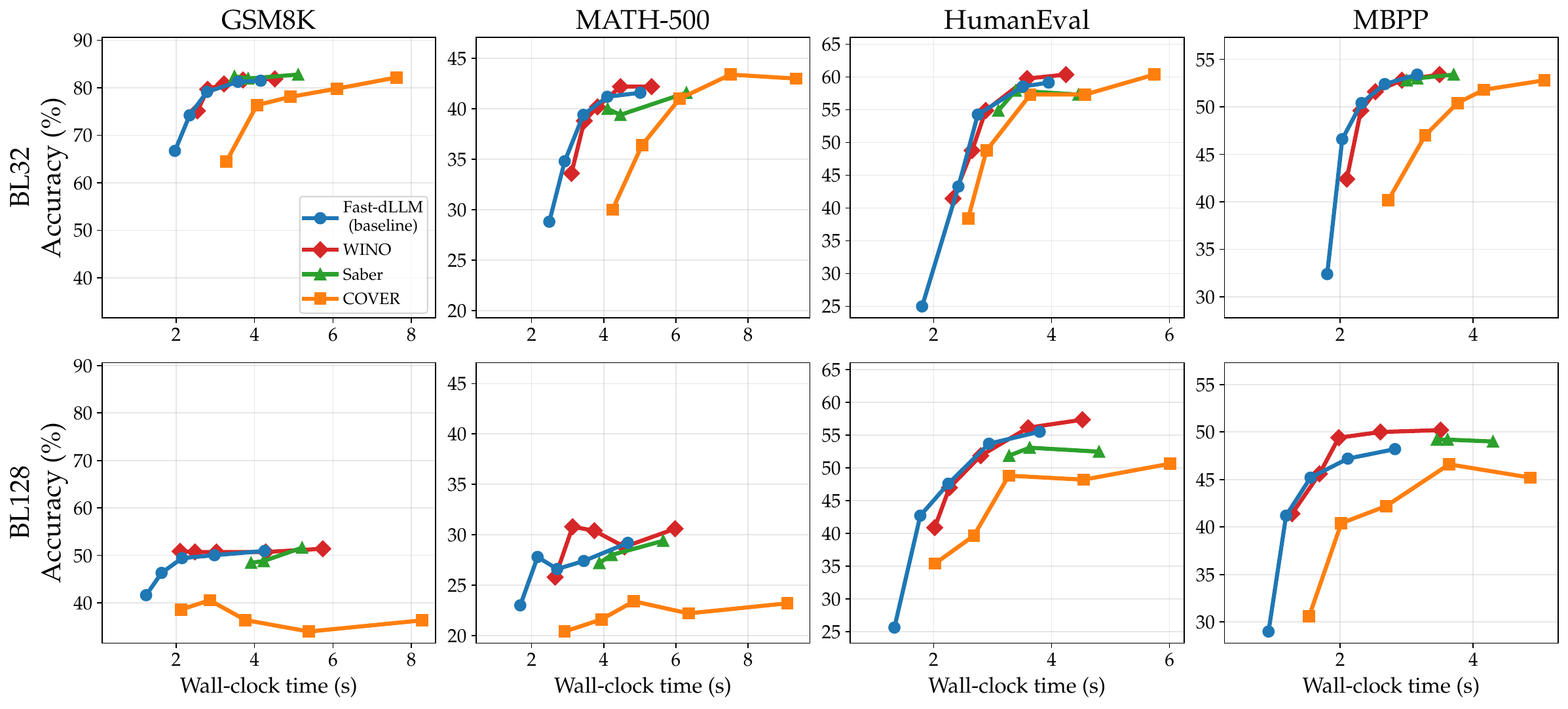}
    \caption{Dream-v0-Instruct-7B}
\end{subfigure}
    \caption{\Cref{fig:wino_greedy,fig:wino_greedy_dream} replicated using latency (wall-clock time) as an efficiency measure instead of network function evaluations (NFEs). Once their wall-clock overhead is taken into account, the gains of remasking diminish further.}
    \label{fig:wallclock}
\end{figure}

\newpage

\FloatBarrier

\section{Additional WINO experiments}
\subsection{Ablations and tuning}
\label{sec:app_ablations}

\begin{figure}[!ht]
\centering
\includegraphics[width=0.7\linewidth]{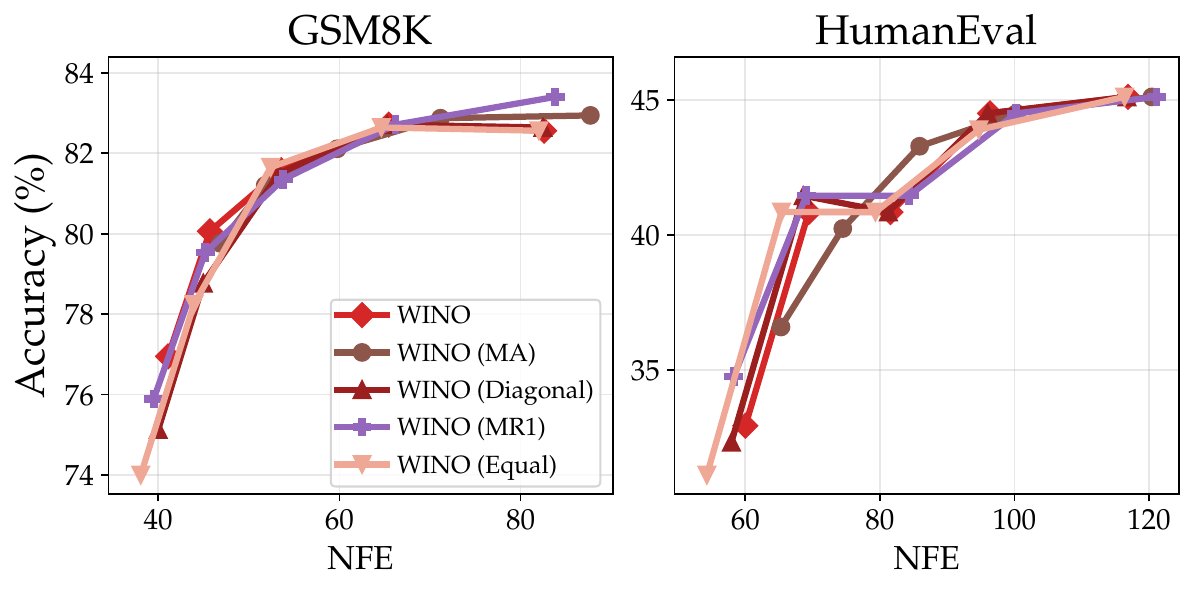}
    \caption{Ablations of four WINO design choices across GSM8K and HumanEval (LLaDA-8B-Instruct, $BL=32$, $\tau=0$). The performance across all variants is similar with no variant showing meaningful gains.}
    \label{fig:ablations1}
\end{figure}

\paragraph{Design choices} To ensure that our findings in \Cref{sec:exp_takeaway1} are robust across various design choices of WINO, we perform the following ablations and report results in \Cref{fig:ablations1}:
\begin{itemize}
    \item \textbf{Consistency-based remasking criterion (\textit{Equal})} Recall that WINO selects positions for remasking via confidence thresholding based on shadow position confidences (Equation~\ref{eq:wino}). To remove the effect of the exact choice of the remasking threshold $\lambda_2$, we instead consider here the consistency-based approach where the position is flagged for remasking if the max token under the shadow predictive distribution is not the same as the current token:
    \begin{align*}
        \mathcal{R}_t := \{ k \in \mathcal{B} \setminus \mathcal{M}_t \mid \arg\max_{v \in \mathcal{V}}q_{\theta}^{k_s}(v \mid \tilde{\vx}_t) \neq x_t^k\} \: .
    \end{align*}
    \item \textbf{Loop-guard (\textit{MR1})} To measure the impact of the exact loop-guard implementation, we consider here an alternative implementation where each position is allowed to be remasked at most once. 
    \item \textbf{Shadow attention mask (\textit{Diagonal})} Instead of using a full attention mask for the shadow region as originally proposed in WINO (\Cref{fig:attention_mask_wino}), we use a diagonal mask which prevents shadow positions from attending to each other. 
    \item \textbf{Max-accept (\textit{MA})} The original WINO implementation imposes a cap on the number of positions that can be unmasked at each step.\footnote{\url{https://github.com/Feng-Hong/WINO-DLLM/blob/main/LLaDA/decoding.py}} Since this cap is neither described in the WINO paper nor part of the Fast-dLLM unmasking procedure \citep{wu2025fastdllmtrainingfreeaccelerationdiffusion}, we disable it for the experiments in the main paper and re-enable it here to study its impact.
\end{itemize}

\begin{figure}[!ht]
\centering
\includegraphics[width=\linewidth]{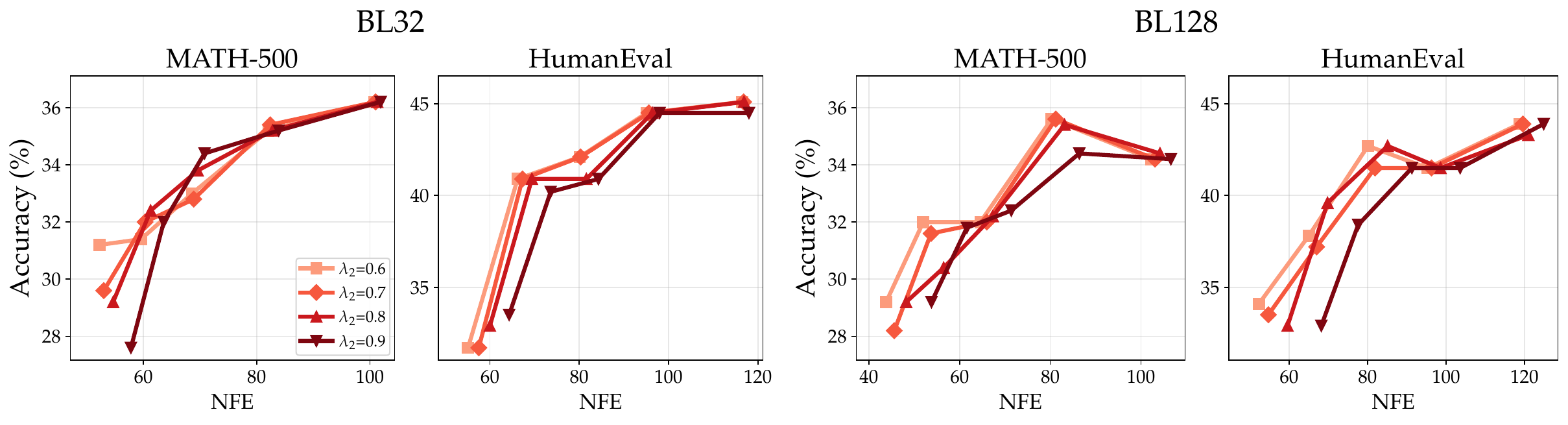}
    \caption{WINO Pareto frontiers for different remasking thresholds $\lambda_2$ (LLaDA-8B-Instruct, $\tau=0$, $BL \in \{32, 128\}$), sweeping the unmasking threshold $\lambda_1 \in \{0.5, 0.6, 0.7, 0.8, 0.9\}$. Results are largely insensitive to the choice of $\lambda_2$.}
    \label{fig:lambda2}
\end{figure}

\paragraph{Remasking threshold} Throughout the main paper, we use the WINO remasking threshold $\lambda_2 = 0.8$, one of the values suggested in the WINO paper \citep{hong2025wide}. To test whether our findings are specific to this choice, we additionally vary $\lambda_2 \in \{0.6, 0.7, 0.8, 0.9\}$ and report results in \Cref{fig:lambda2}. The Pareto frontiers are largely insensitive to $\lambda_2$, and WINO's improvement over Fast-dLLM's top performance at $BL=32$ remains marginal for every threshold (${\sim}0.5\%$ for $\lambda_2 \in \{0.6, 0.7, 0.8\}$ and ${\sim}0.2\%$ for $\lambda_2 = 0.9$, averaged over MATH-500 and HumanEval).

\subsection{Shadow-token consistency}
\label{sec:app_consistency}

\begin{figure}[!htbp]
\centering
\includegraphics[width=0.4\linewidth]{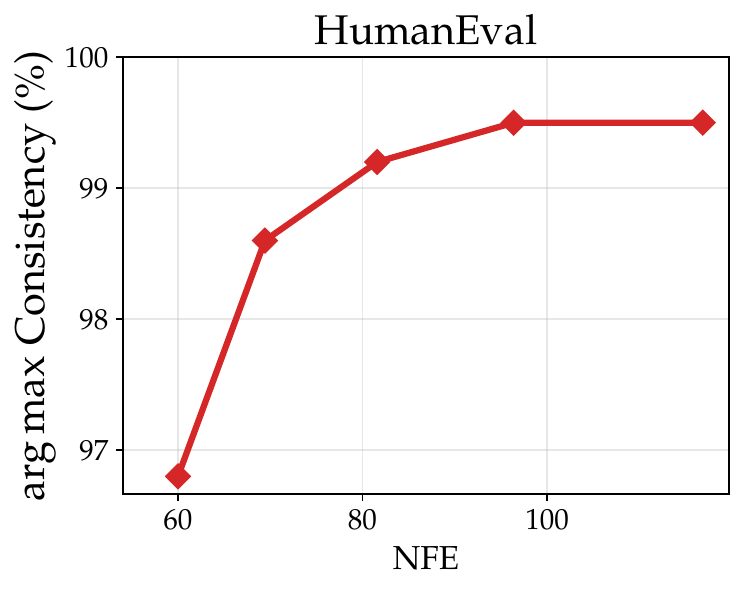}
    \caption{Consistency rate ($\{\arg\max_vq_{\theta}^{k_s}(v \mid \tilde{\vx}_t) = \arg \max_v q_{\theta}^k(v \mid \vx_{t, -k})\}$) of the highest-confidence tokens between the shadow and oracle predicted tokens \emph{without} remasking ($\lambda_2=0$, LLaDA-8B-Instruct, HumanEval, $BL=32$). Unmasking fewer tokens per step, i.e., increasing the threshold, further improves shadow predictions, potentially due to fewer approximations of KV matrices of other tokens at each step.}
    \label{fig:disagreement}
\end{figure}

To further validate the shadow-token approximation (\Cref{sec:exp_takeaway2}), we measure how often the shadow and oracle (leave-one-out) predictions agree on the top-1 token. As shown in \Cref{fig:disagreement}, the consistency rate is above 96\% for all unmasking thresholds $\lambda_1$, and increases as fewer tokens are unmasked per step.

\subsection{Flip-flop analysis}
\label{sec:app_flipflop}

\paragraph{Expanded remasking}
\label{sec:app_expand_remask}
To better understand the high flip-flop rates observed in \Cref{fig:flipflop}, we perform additional ablations that vary the set of positions remasked at each sampling step. Beyond the position $k$ flagged by confidence thresholding (Equation~\ref{eq:wino}), we additionally remask either its \emph{spatial} or \emph{temporal} neighbors. Formally, each flagged position $k \in \mathcal{R}_t$ is remasked together with its spatial or temporal neighborhood:
\begin{align*}
    \mathcal{N}^{S}(k) &:= \{\, j \in \mathcal{B} \setminus \mathcal{M}_t \mid |j - k| \le S \,\}, \\
    \mathcal{N}^{T}(k) &:= \{\, j \in \mathcal{B} \setminus \mathcal{M}_t \mid 0 \le u_j - u_k \le T - 1 \,\},
\end{align*}
where $u_j$ is the step at which position $j$ was last unmasked. Spatial neighbors are thus the $S$ positions on either side of $k$, and temporal neighbors span the $T$ steps from $k$'s own unmasking onward ($T = 1$ the same step, $T = 2$ that step and the next). We denote the resulting variants WINO (S1), WINO (S2), WINO (T1), and WINO (T2), where the number gives the neighborhood size $S$ or $T$. In addition, to ensure convergence, we switch to the aforementioned \textit{MR1} loop-guard, in which each position can be flagged for remasking only once, but can still be remasked as a neighbor of other positions.

\begin{figure}[!htbp]
\centering
\includegraphics[width=\linewidth]{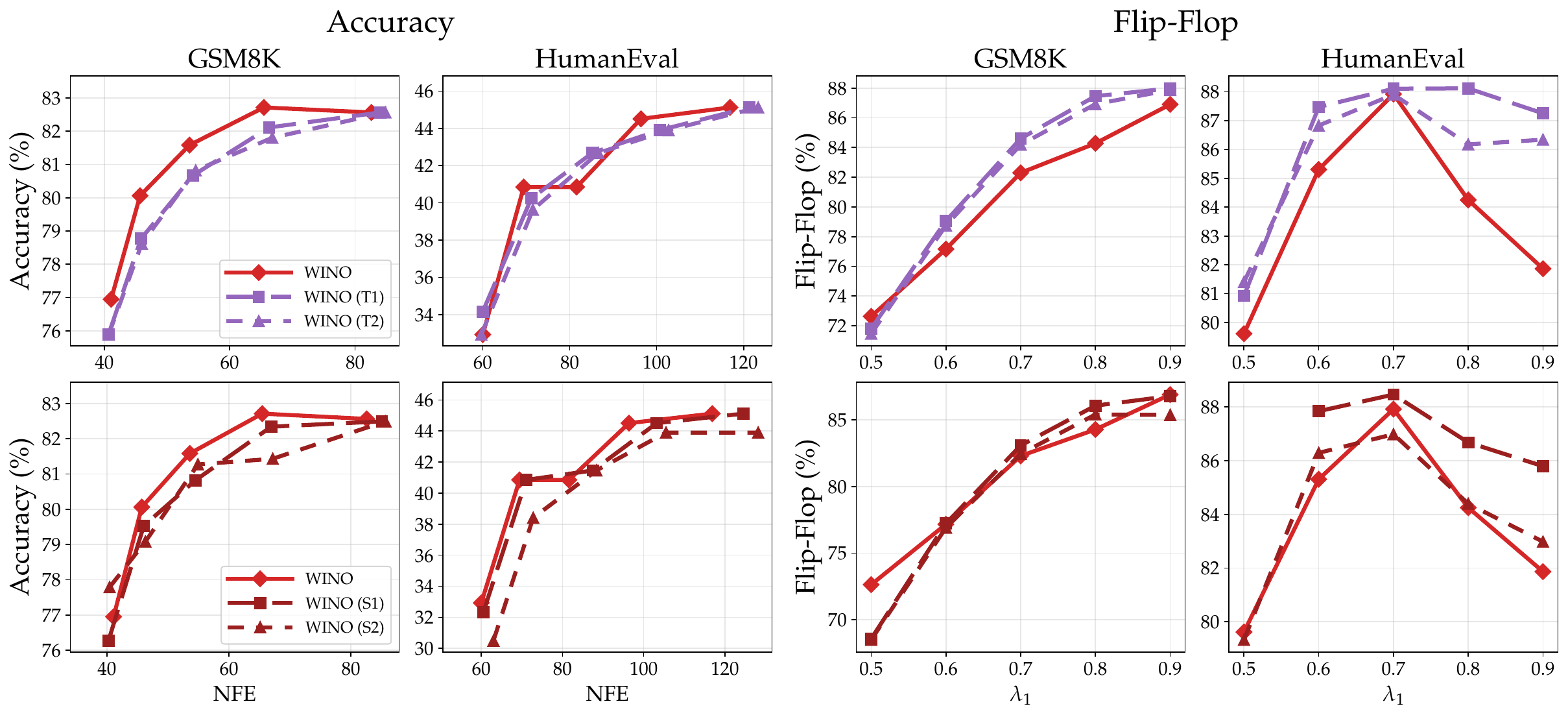}
    \caption{Accuracy (left, vs.\ NFE) and flip-flop rate (right, vs.\ $\lambda_1$) of expanded remasking ablations that additionally remask temporal (top) or spatial (bottom) neighbors (LLaDA-8B-Instruct, GSM8K and HumanEval, $BL=32$).}
    \label{fig:expansion}
\end{figure}

As shown in \Cref{fig:expansion} (left), neither expansion method improves over default WINO: both variants match or \emph{degrade} accuracy across the NFE range, with spatial expansion degrading more as the neighborhood grows ($S = 2$ vs.\ $S = 1$). In addition, \Cref{fig:expansion} (right) shows that expanded remasking does not reduce the flip-flop rate, and often increases it. This argues against the cascading-dependency explanation: even when the surrounding context that would supposedly pin the prediction at $x_t^k$ is partially removed, the dLLM still re-proposes the same token. The bottleneck therefore appears to lie in the model's predictive distribution at position $k$ itself, rather than in the conditioning structure of the unmasked context.

\begin{figure}[!htbp]
\centering
\includegraphics[width=\linewidth]{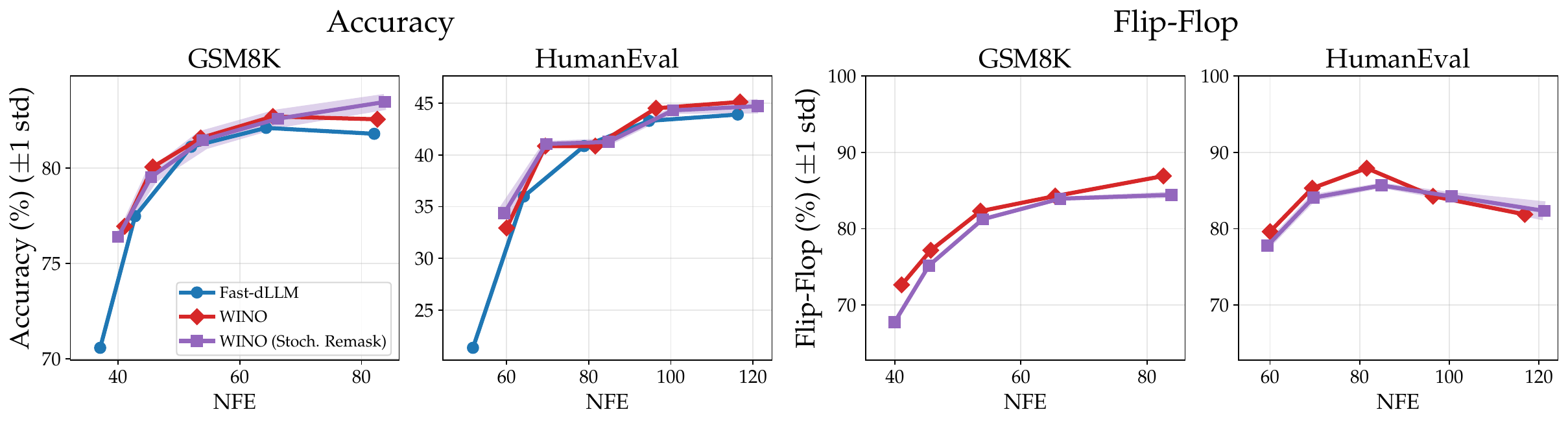}
    \caption{Accuracy (left) and flip-flop rate (right) of WINO with greedy ($\tau=0$) vs.\ stochastic ($\tau=1$) sampling at remasked positions (LLaDA-8B-Instruct, GSM8K and HumanEval, $\lambda_2 = 0.8$, $BL=32$), sweeping the unmasking threshold $\lambda_1 \in \{0.5, 0.6, 0.7, 0.8, 0.9\}$. Fast-dLLM is shown for reference. Stochastic remasking results are averaged over 3 seeds ($\pm 1$ std).}
    \label{fig:stoch_remask}
\end{figure}

\paragraph{Stochastic sampling at remasked positions}
\label{sec:app_stoch_remask}
Another possible explanation for the high flip-flop rates is greedy decoding ($\tau=0$): when a remasked position is selected for unmasking again, the greedy sampler deterministically reproduces the model's most likely token, which is often the token that was just remasked. To test this, we sample remasked positions with $\tau=1$ whenever they are selected for unmasking again, while all other positions are still decoded greedily, encouraging the model to explore alternative tokens rather than reproduce its earlier prediction. As shown in \Cref{fig:stoch_remask}, stochastic sampling at remasked positions has no significant effect on accuracy and only slightly lowers the flip-flop rates.

\FloatBarrier
\paragraph{Correct vs.\ incorrect generations}
\label{sec:app_ff_correctness}
To test whether flip-flops concentrate on generations where the model makes mistakes, we split the flip-flop rate by whether the final answer is correct (LLaDA-8B-Instruct, WINO with $\lambda_1 = 0.6$, $\lambda_2 = 0.8$, $BL=32$). As shown in \Cref{tab:ff_correctness}, the rates are nearly identical for correct and incorrect generations across all four datasets, suggesting that flip-flops cannot reliably indicate when the model is likely to produce an incorrect answer.

\begin{table}[H]
\centering
\caption{Average flip-flop rate (\%) for correct and incorrect generations (LLaDA-8B-Instruct, WINO, $\lambda_1 = 0.6$, $\lambda_2 = 0.8$, $BL=32$). Number of generations in parentheses.}
\label{tab:ff_correctness}
\begin{tabular}{lcccc}
\toprule
 & GSM8K & MATH-500 & HumanEval & MBPP \\
\midrule
Correct   & 74.6 (1056) & 76.3 (162) & 85.6 (67) & 81.9 (162) \\
Incorrect & 77.9 (263)  & 77.6 (338) & 82.4 (97) & 81.9 (338) \\
\bottomrule
\end{tabular}
\end{table}

\subsection{Token entropy under stochastic decoding}
\label{sec:app_entropy}
To better understand the loss in diversity of WINO under stochastic decoding (\Cref{fig:wino_passatk}), we measure the entropy of the model's predictive distribution at each position when it is unmasked (LLaDA-8B-Instruct, $BL=32$, $\tau=0.8$, $\lambda_1 = 0.6$, $\lambda_2 = 0.8$, 64 samples per problem). Since WINO can unmask a position more than once, we report the entropy at both its first and final unmasking. As shown in \Cref{tab:entropy}, WINO remasks only ${\sim}4\%$ of positions, but these are highly uncertain when first unmasked. After remasking, they are re-committed with much lower entropy, below that of the average position, which lowers WINO's overall token entropy compared to Fast-dLLM.

\begin{table}[H]
\centering
\caption{Mean token entropy at the first and final unmasking of each position (LLaDA-8B-Instruct, $BL=32$, $\tau=0.8$, $\lambda_1 = 0.6$, $\lambda_2 = 0.8$, 64 samples per problem). Since Fast-dLLM unmasks each position once, its final entropy equals its first ($=$). WINO remasks 3.6\% (MATH-500) and 4.4\% (HumanEval) of positions.}
\label{tab:entropy}
\begin{tabular}{llcccc}
\toprule
 & & \multicolumn{2}{c}{MATH-500} & \multicolumn{2}{c}{HumanEval} \\
\cmidrule(lr){3-4} \cmidrule(lr){5-6}
Method & Positions & First & Final & First & Final \\
\midrule
Fast-dLLM & All & 0.64 & $=$ & 0.79 & $=$ \\
WINO & All & 0.63 & 0.61 & 0.77 & 0.74 \\
WINO & Remasked & 1.15 & 0.43 & 1.38 & 0.55 \\
\bottomrule
\end{tabular}
\end{table}

\clearpage

\end{document}